\documentclass[runningheads]{llncs}

\usepackage{eccv}

\usepackage{eccvabbrv}

\usepackage{graphicx}
\usepackage{booktabs}
\usepackage{multirow}
\usepackage{graphicx}
\usepackage{amsmath}
\usepackage{amssymb}
\usepackage{booktabs}
\usepackage{xcolor}
\usepackage{bm}
\usepackage{pifont}
\usepackage{enumitem}
\usepackage{soul}
\usepackage{colortbl}
\usepackage{dblfloatfix}
\usepackage{siunitx}
\usepackage{balance}
\usepackage{rotating}
\usepackage{makecell}
\usepackage{caption}
\usepackage{multicol}

\usepackage{wrapfig}

\usepackage[accsupp]{axessibility}  

\usepackage{hyperref}

\usepackage{orcidlink}

\newcommand{\tit}[1]{\smallbreak\noindent\textbf{#1.}}

\newcommand{\cmark}{\ding{51}}%

\begin{document}

\title{\textit{\textmu}gat: Improving Single-Page Document Parsing by Providing Multi-Page Context} 
\titlerunning{\textit{\textmu}gat: Improving Document Parsing with Multi-Page Context}

\author{Fabio Quattrini$^*$ \and Carmine Zaccagnino$^*$ \and Silvia Cascianelli \and \\ Laura Righi \and Rita Cucchiara}
\authorrunning{F.~Quattrini et al.}
\institute{University of Modena and Reggio Emilia, Modena, Italy\\
\email{\{name.surname\}@unimore.it}\\
$^*$Equal Contribution}

\maketitle

\begin{abstract}
    Regesta are catalogs of summaries of other documents and, in some cases, are the only source of information about the content of such full-length documents. For this reason, they are of great interest to scholars in many social and humanities fields. In this work, we focus on Regesta Pontificum Romanum, a large collection of papal registers. Regesta are visually rich documents, where the layout is as important as the text content to convey the contained information through the structure, and are inherently multi-page documents. Among Digital Humanities techniques that can help scholars efficiently exploit regesta and other documental sources in the form of scanned documents, Document Parsing has emerged as a task to process document images and convert them into machine-readable structured representations, usually markup language. However, current models focus on scientific and business documents, and most of them consider only single-paged documents. To overcome this limitation, in this work, we propose \textit{\textmu}gat, an extension of the recently proposed Document parsing Nougat architecture, which can handle elements spanning over the single page limits. Specifically, we adapt Nougat to process a larger, multi-page context, consisting of the previous and the following page, while parsing the current page. Experimental results, both qualitative and quantitative, demonstrate the effectiveness of our proposed approach also in the case of the challenging Regesta Pontificum Romanorum.
  
  \keywords{Digital Humanities \and Document Parsing \and Optical Character Recognition}
\end{abstract}

\section{Introduction}
\label{sec:intro}

\begin{figure}[t]
    \centering
    \setlength{\tabcolsep}{0.1pt}
    \begin{tabular}{ccc}
        \includegraphics[width=0.33\linewidth]{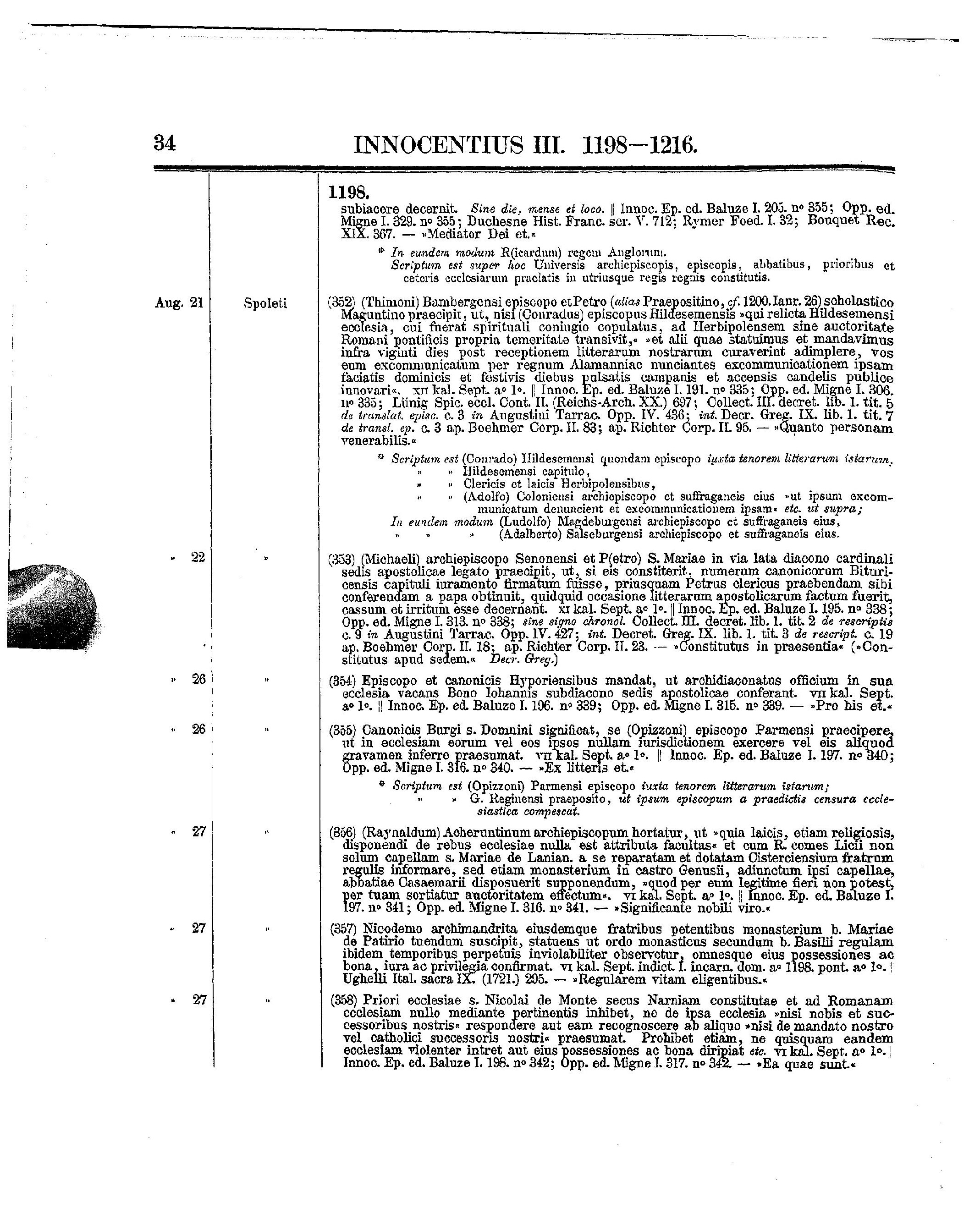}&
        \includegraphics[width=0.33\linewidth]{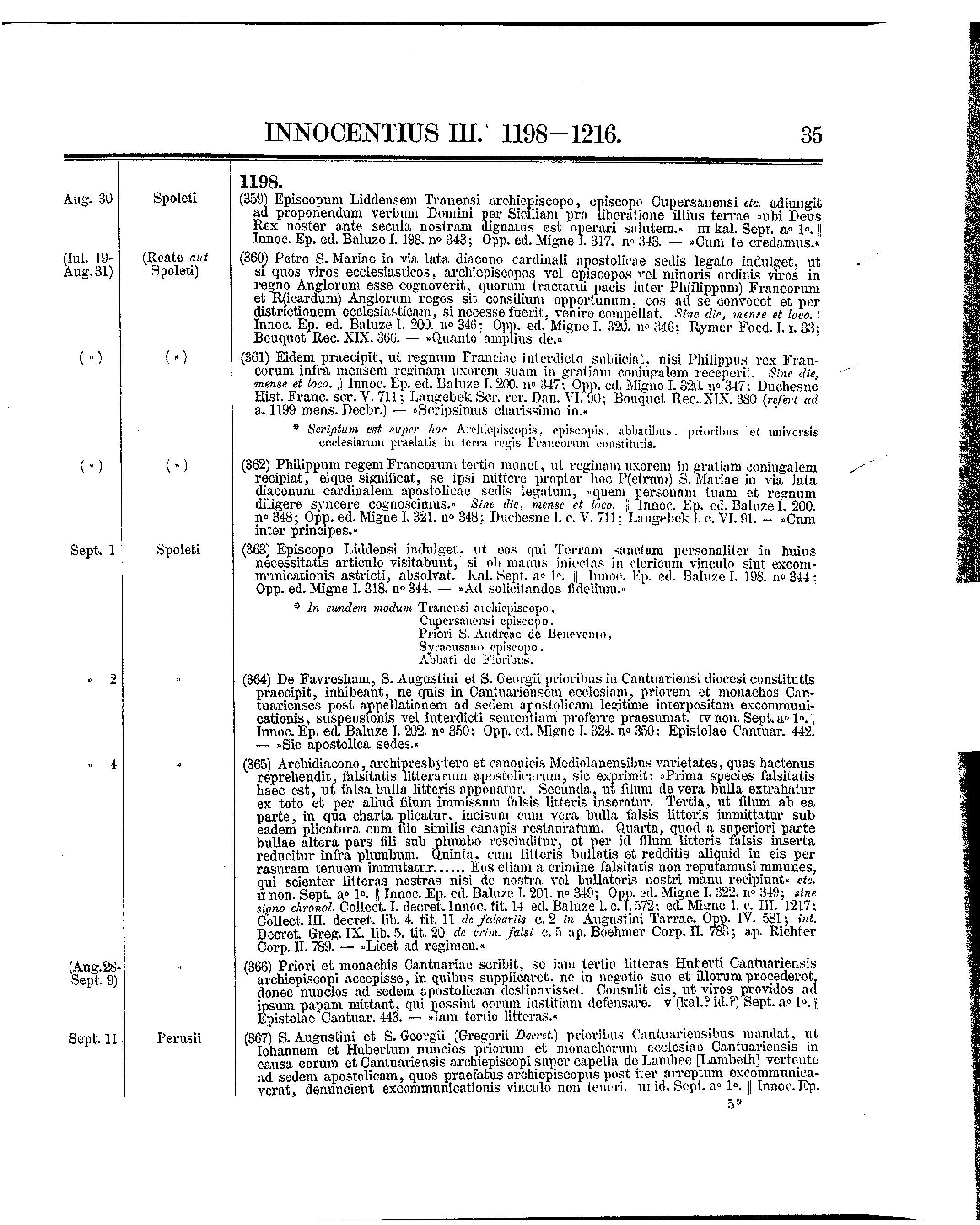}&
        \includegraphics[width=0.33\linewidth]{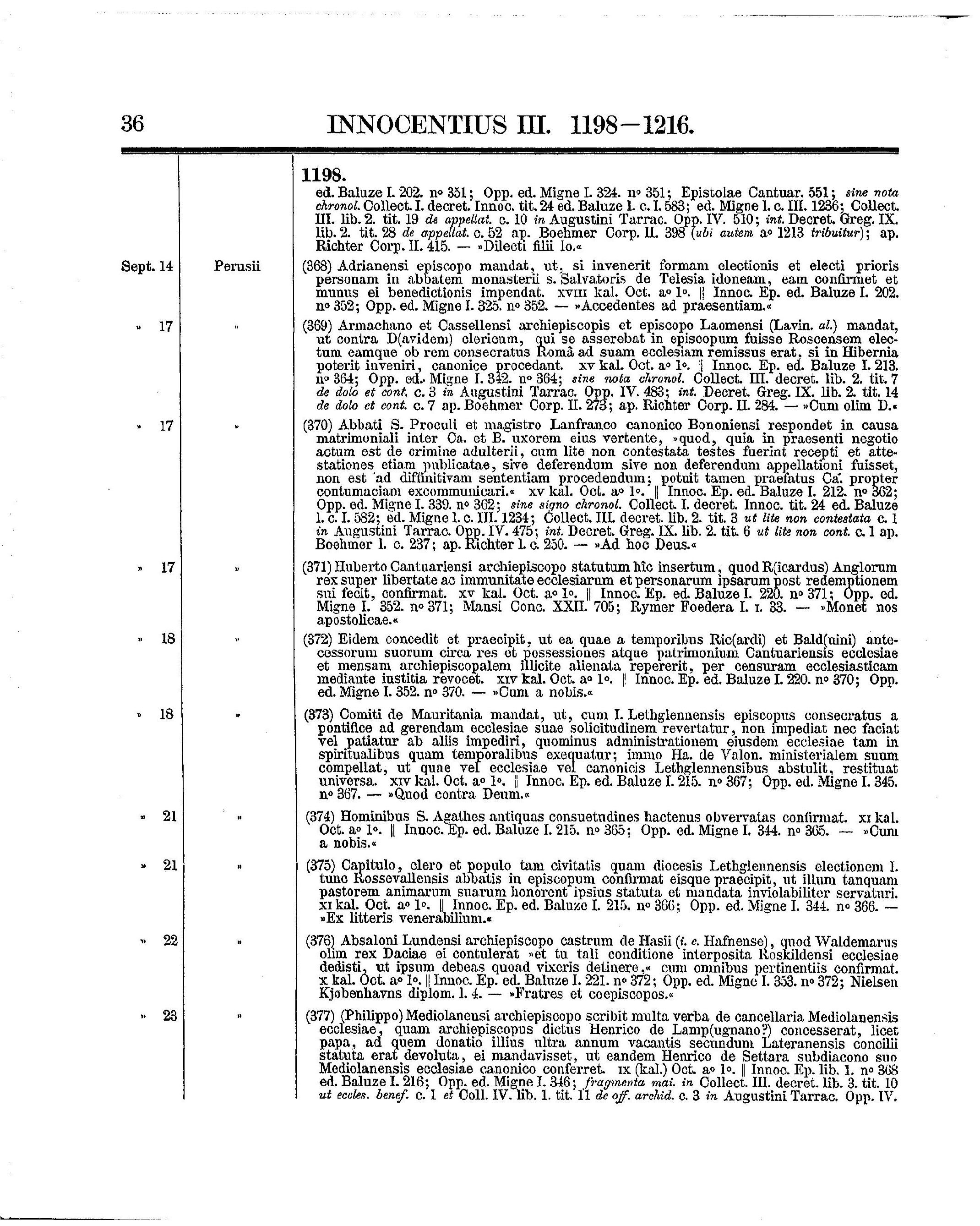}
    \end{tabular}
    \caption{Three consecutive pages from the Regesta Pontificum Romanorum: summaries are both simple-layout paragraphs and text in a tabular layout, and the corresponding information about date and place is given via layout. Such entries (summaries and metadata) span multiple pages, with interleaved information and layout elements.}
    \label{fig:overview}
\end{figure}

A vast majority of Cultural Heritage and historical data comes in the form of digitized document images. 
Among those, \textit{regesta} collections are a valuable source of information for scholars of many different fields of humanities and social sciences (\eg~history, religious studies, linguistic and literary studies, \etc). In fact, regesta serve both as organized collections of sources and data related to a vast range of topics, places, and individuals and as indexing tools for scholars to find the historical documents preserved by archives all around the world. 
Specifically, regesta are chronologically ordered catalogs of summaries of historical documents, which, in some cases, are the only existing source of information about the full-length documents they summarize. A regestum is a highly structured summary of a document, both visually and textually. Such documents have been created by following a precise and consolidated methodology, which involves the extraction of information from specific sections of the full-length document and its translation and codification into a language shared by the community of scholars at the time the regestum has been edited. This methodology to summarize historical documents has been specifically formalized and used to collect, index, and summarize medieval documents. 
Regesta collections have been compiled by intellectuals and scholars starting from the 19th century until the present, creating a rich corpus of historical data. However, only scanned versions of the regesta collections are available, not in machine-readable format. This is because this wealth of historical data is preserved in printed texts from the 19th and 20th centuries, with fonts and layouts that are difficult to process with traditional OCR systems. Specifically, regesta are characterized by difficult-to-read ancient fonts, extensive use of abbreviations and acronyms, and, peculiarly, a multi-column layout (different from edition to edition). In fact, regesta are documents whose layout is as important as the content since it helps to correctly correlate the contained information through the structure. In this sense, \textit{regesta are visually rich documents}. 

Visually-rich Document Understanding (VrDU) is an area of research that aims to analyze and extract information from document images. Unlike other vision-based data, the text is an integral part of the image. Therefore, to effectively perform this task, it is necessary to consider and combine textual, visual, and spatial information. Given the huge range of different layouts, conservation status, scan quality, and content of documents, existing approaches usually entail pertaining models to perform tasks like parsing and masked language/visual modeling~\cite{xu2020layoutlm, xu2020layoutlmv2, huang2022layoutlmv3, kim2022ocr, davis2022end, li2022dit, blecher2023nougat} on large-scale general datasets like IIT-CDIP~\cite{lewis2006building} or CC-PDF~\cite{turski2023ccpdf}. Then, these models are fine-tuned on specific datasets for different subtasks, such as Document Classification~\cite{lewis2006building, harley2015evaluation}, Document Layout Analysis~\cite{zhong2019publaynet, li2020docbank}, Key Information Extraction~\cite{jaume2019funsd}, and Document Question Answering (DQA)~\cite{mathew2021docvqa}. Despite being an active research area, there are still many challenges and open questions concerning scalability, generalization, and sub-task definitions~\cite{van2023document, van2024beyond}. 
One aspect that is receiving increasing research interest is that in most settings (and relative datasets), the considered documents are single-paged or seen as an independent set of single pages. This limitation severely hinders the capabilities of the resulting developed models in real-world scenarios, both in terms of performance( as they cannot leverage multi-page information and wrongly consider each page as independent) and in terms of possible applications~\cite{van2023document,van2024beyond}. In this respect, some recent works have tackled this problem by also proposing new multi-page document datasets for VrDU~\cite{van2023document, liu2024focus}, Classification~\cite{van2024beyond}, and DQA~\cite{van2023document}. In particular, multi-page DQA is attracting interest, and some approaches have been proposed that extend single-page pretrained architectures to handle multi-paged inputs~\cite{tito2023hierarchical, blau2024gram}.

Particularly relevant in the context of VrDU is the Document Parsing task, which consists in extracting a structured representation of a document in markup language. This task is considered both for pre-training~\cite{davis2022end, kim2022ocr} and as the final goal~\cite{blecher2023nougat, sun2024locr, lee2023pix2struct, lv2023kosmos} in VrDU. 
Academic documents are particularly interesting for the VrDU community~\cite{blecher2023nougat, sun2024locr} because of the challenges they pose, which make them a valuable tool for developing general and effective VrDU models. 
Most of these documents are images in PDF format, obtained from scanned documents produced before the age of the Internet, which usually present complicated and highly variable layouts with tables, formulas, and graphs intertwined with text. Moreover, these documents are almost always multi-page, with text split between pages and interleaved with images or tables and cross-page references.
However, State-of-the-art works are trained to parse single pages. 
We argue that, due to the multi-page nature of most visually-rich documents,
when parsing a page, the models can benefit from taking into account the \textit{context} around the page, which we define as the previous and next page. 

In this work, we treat academic document parsing as a proxy task for obtaining a model able to parse regesta collections without fine-tuning on such documents. In fact, due to their information-carrying layout and the underrepresented language in which they are written (medieval Latin), obtaining a large amount of high-quality annotated regesta pages requires multidisciplinary skills and, therefore, is costly and time-consuming. Thus, a language-agnostic, multi-page document parsing model that can be used off-the-shelf on regesta would be highly valuable for the Digital Humanities community. For this reason, in this work, we tackle multi-page document parsing by proposing an adaptation strategy to make pretrained OCR-free single-page document parsing models able to handle multiple pages as context for increasing the consistency of the parsing output. In particular, we consider the recently proposed Nougat~\cite{blecher2023nougat} architecture as the base single-page document parsing model. We call our approach $\mu$\textit{gat}, which stands for \textit{multi-page Nougat}.

As a use-case, we consider the \textit{Regesta Pontificum Romanorum}~\cite{potthast1874regesta}, edited by August Potthast between 1874 and 1895 (see some pages in~\cref{fig:overview}). This is a well-known collection of summaries of documents produced and issued by popes and pontifical chanceries (papal letters, bulls, decretals, \etc) from the origins of the Latin Church to the beginning of the 14th Century. By providing detailed records of papal decisions, letters, and administrative actions, they serve as a scholarly resource for the study of the Latin Church and, more broadly, for the history of ancient and medieval Europe. 
The Regesta Pontificum Romanorum can be seen as a multi-page, three-column table where each page has a header bearing the pontiff who authored the summarized documents and the year. The table entries are relative to different full-length documents and report the following information: 1) the date (month and day) in which the document has been edited; 2) the place where it was edited; and 3) the text of the regestum, \ie~the summary. The regestum itself has a specific structure and contains the following information in a fixed order: addressee; summary of the contents; date according to the ancient calendar; references to find the document in the original version (on parchment) or in a printed edition, and the incipit of the text. The loss of any of these elements or a reversal of their order would make the summary incomprehensible and the original medieval document impossible for scholars to find. Therefore, for its relevance for humanities studies and for the challenges posed for VrDU, we choose this collection as our use-case to validate the proposed approach for multi-page context-enhanced document parsing.

To foster the research in multi-page document parsing, we release our code for both the dataset preparation and the proposed approach at \url{https://github.com/aimagelab/mugat}.
\section{Related Work}
\label{sec:related}
\tit{VrDU Datasets}
The field of VrDU encompasses both general and task-specific datasets~\cite{lewis2006building, turski2023ccpdf, lewis2006building, harley2015evaluation, zhong2019publaynet, li2020docbank, jaume2019funsd, mathew2021docvqa, van2023document, van2024beyond}. The large scale of these datasets and the emergence of the Transformer~\cite{vaswani2017attention} has revolutionized Document AI~\cite{xu2020layoutlm, xu2020layoutlmv2, huang2022layoutlmv3, kim2022ocr, davis2022end, li2022dit, blecher2023nougat, tang2023unifying, appalaraju2021docformer, appalaraju2024docformerv2, lee2023pix2struct, lv2023kosmos}. However, these benchmarks have been shown to suffer from single-domain scope, limited document types, and single-page samples~\cite{van2023document, van2024beyond}. With MP-DOCVQA~\cite{tito2023hierarchical}, DUDE~\cite{van2023document}, and $\text{RVL-CDIP}_\text{MP}$~\cite{van2024beyond}, the research community has started to consider the multi-page case for sub-tasks like (Document) Visual Question Answering and Classification. In this work, we focus on parsing, which refers to translating a document's image into a structured markup representation~\cite{blecher2023nougat}. 
In~\cite{blecher2023nougat}, most of the training documents are parsed from arXiv. In particular, the authors process the source files by using \LaTeX ML\footnote{\url{dlmf.nist.gov/LaTeXML/}} to standardize and convert them into HTML5. Then, they convert the HTML files into Markdown and split the pages by exploiting heuristics and fuzzy matching. However, this procedure results in independent single-page samples, and the parsing model trained on such data is only capable of handling single pages without considering that some elements can span multiple pages. To overcome this limitation, in this work, we exploit a specifically-collected dataset of academic documents containing both contiguous and sparse pages to allow the training and benchmarking of flexible models capable of managing the context when provided.

\tit{VrDU Methods}
VrDU has attracted increasing interest in recent years and is dominated by Transformer-based models~\cite{vaswani2017attention}. Methods can be categorized whether they include an external OCR input or not. In the former category, the LayoutLM family~\cite{xu2020layoutlm, xu2020layoutlmv2, huang2022layoutlmv3} is based on the Transformer-Encoder BERT~\cite{devlin2018bert} and performs masked layout prediction. In UDOP~\cite{tang2023unifying} and DocFormer~\cite{appalaraju2021docformer, appalaraju2024docformerv2}, Transformers are trained to perform several pre-training tasks and align the textual and visual information. 
On the other hand, OCR-free methods have recently shown that competitive results can be achieved without using an external OCR~\cite{davis2022end, kim2022ocr}, which also increases generalization to unseen languages and domains. In parsing, the recent Nougat approach~\cite{blecher2023nougat} has achieved state-of-the-art performance on academic document parsing. This architecture is based on a Swin Encoder~\cite{liu2021swin} that takes as input single-page images and outputs embedding vectors. These are then processed by a mBART~\cite{lewis2019bart} decoder that autoregressively outputs the Markdown code by using the same tokenizer as~\cite{taylor2022galactica}. 
Note that all the approaches mentioned above, including Nougat, entail considering documents as single pages. This is often not the case in real-world applications, where documents exhibit parts (\eg~paragraphs, references, or footnotes) that can sit between multiple pages. Therefore, in our proposed approach, we adopt the Nougat architecture and extend it to process multi-page inputs to increase the consistency of its output.

\tit{Adapting single-page models to multi-page scenarios} 
Some previous works have proposed approaches to extend pretrained VrDU models working on single pages for handling multi-page inputs, especially for the DQA task (with the multi-page DocVQA variant of the task~\cite{kang2024multi}). 
For example, in Hi-VT5~\cite{tito2023hierarchical}, the authors fine-tuned the Document Image Transformer model~\cite{li2022dit} to encode the input the single-page images into multiple global tokens, which are then fed to a T5~\cite{raffel2020exploring} model, trained to align layout and text representations~\cite{biten2022latr}. However, such global tokens contain too coarse information, which can be sufficient for VQA but would be unsatisfactory for document parsing since this latter task requires detailed representations. 
Another example is GRAM~\cite{blau2024gram}, in which a single-page DocFormerv2~\cite{appalaraju2024docformerv2} encoder is extended to use multiple pages as input by inserting global tokens in the page representation and alternating attention operations over the single pages and the global tokens. 
In this work, we take inspiration from the Perceiver~\cite{jaegle2021perceiver} model and implement an adapter that extracts relevant information from the previous and next pages and inserts them into learnable tokens, which are then concatenated to the representation of the current page so that the decoder can parse it while keeping into consideration some information from the context. 
\section{Proposed Approach}
\label{sec:method}
In this section, we first provide the details about the Document parsing architecture that we use as the base model and the Adapter model we devise to enhance its performance by exploiting multiple pages as context when parsing the page of interest. Then, we describe the semi-automatic procedure we adopt to collect increasingly challenging data used for training and validating our model. These include complex-layout academic papers, long tables, and pages from the Regesta Pontificum Romanorum.

\begin{figure}[t]
    \centering
    \includegraphics[width=\linewidth]{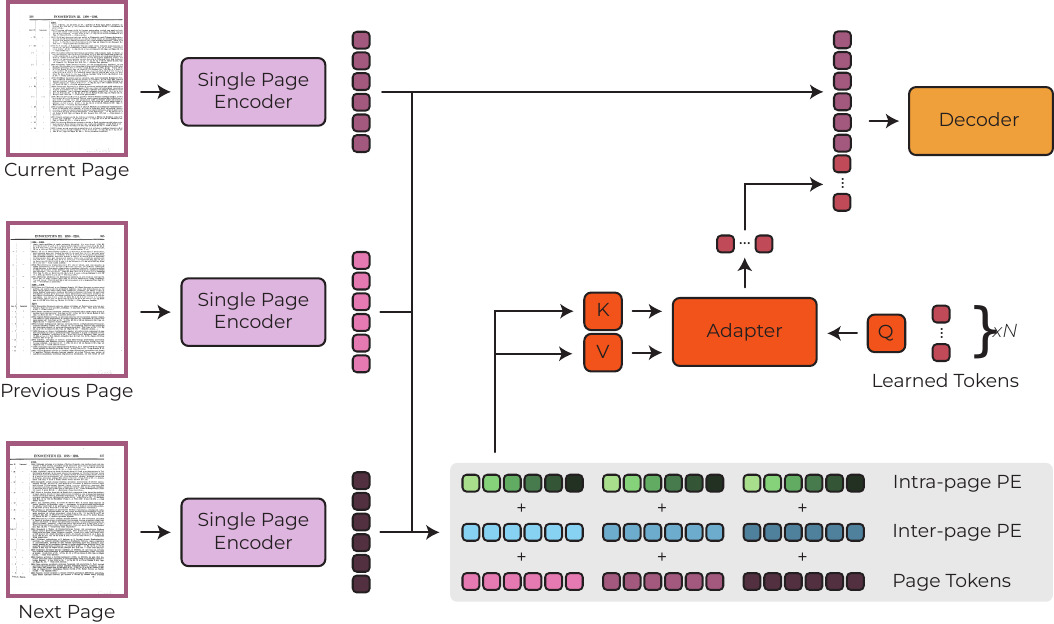}
    \caption{Our proposed $\mu$gat architecture. The Visual Encoder embeds the previous, current, and next page. These embeddings are then concatenated and, after the addition of inter-page and inter-page positional embeddings, fed into the key and value of the adapter's cross-attention layers, which extract information into the learned $N$ tokens representing compressed information about the pages. These tokens are then appended to the current page tokens and fed to the auto-regressive Decoder.}
    \label{fig:mugat-architecture}
\end{figure}

\subsection{\textit{\textmu}gat Model}
\tit{Base Architecture}
Our approach is based on the idea of introducing multi-page context as input to encoder-decoder document parsing models to increase the consistency of their output by exploiting information from multiple pages. In this work, we build upon the recently proposed single-page parsing Nougat~\cite{blecher2023nougat} architecture. This is an Encoder-Decoder Transformer~\cite{vaswani2017attention}, inspired by Donut~\cite{kim2022ocr} and specialized in document parsing. In particular, the encoder is a Visual Encoder implemented as a Swin Transformer~\cite{liu2021swin}, which takes as input a document image $\mathbf{x} \in \mathbb{R}^{H_0 \times W_0}$ and embeds it into a sequence of $P$ vectors $\mathbf{z} \in \mathbb{R}^{d}$. The decoder is implemented as a mBART\cite{lewis2019bart}, which auto-regressively generates a sequence of tokens by attending the embedded image via cross-attention and its generated tokens via self-attention. The output tokens are then projected to the same vocabulary as the one used in~\cite{taylor2022galactica}. The model is trained to output parsed documents formatted in Markdown, with additional \LaTeX~syntax to represent mathematical equations and have greater flexibility when representing tables.

In this work, we introduce an Adapter module between the Encoder and the Decoder. In particular, we propose to keep the Nougat Encoder frozen and train the Adapter while also fine-tuning the Nougat Decoder. This reduces the resources needed to extend the parsing model, both in terms of training time and annotated data required. 
In particular, the proposed Adapter takes as input the Encoder embeddings of three consecutive pages and extracts relevant information into $N$ learned tokens using the cross-attention mechanism. These tokens are then passed to the Decoder alongside the $P$ embedding vectors extracted by the Encoder for the current page that is being parsed. We provide an overview of our pipeline in~\cref{fig:mugat-architecture} and give the details of the Adapter module in the following.

\tit{Adapter Module}
In order to use multiple pages as input for parsing, a simple approach would be to concatenate multiple page embeddings and feed them directly to the Decoder~\cite{van2023document}. However, our Decoder, as well as most of those used in document parsing solutions, feature attention operators, which scale quadratically with the input size. Therefore, this na\"ive solution would lead to a significant computation overhead. In light of this, we propose to use a dedicated Adapter module. We implement the adapter as a Perceiver~\cite{jaegle2021perceiver} architecture, which takes as input the Encoder outputs for the previous, current, and next pages and consists of a series of $L$ layers, in which cross-attention and feed-forward operations are performed. We train this adapter to generate $N$ $d$-dimensional tokens that summarize information on the content of neighboring pages, which is useful for parsing the current page. 

Specifically, we run the Visual Encoder for each of the three input pages and concatenate the resulting embeddings. Then, we sum a learnable inter-page positional encoding, whose values are the same for all the tokens referring to the same page, and a learnable intra-page positional encoding, whose values are the same for the tokens in the same position of their respective page. The result is linearly projected to the dimension of the keys and values for cross-attention, which is performed using as queries a linear projection of the $N$ learnable tokens. After the cross-attention, the $N$ output tokens go through a position-wise feed-forward neural network, as in the standard Transformer. We propose to maintain the number of learned tokens much smaller than the number of Encoder embedding vectors in order to limit the computation overhead. In fact, the cross-attention operations inside the Adapter scale linearly with the number of context pages considered and have only $N$ additional elements. Moreover, the overhead is limited also in the Decoder since its input is increased by only $N<<P$ tokens. Finally, note that when the previous or the next pages are not available (which can be the case for incomplete documents or for the first and the last pages), we feed the adapter with the embeddings of an empty page. 

\subsection{Training and Fine-tuning Data}\label{sec:method_data}
\tit{Academic Papers}
Previous works tackling document parsing~\cite{blecher2023nougat} usually collect paired single-page images and corresponding markup transcription training data by crawling academic papers PDFs and the associated \LaTeX~source files, from which the ground truth markup parsing can be obtained. However, the authors of these works do not release the data they used. Moreover, the multi-page focus of this work requires preserving the information about the order of the pages to allow feeding our model with the context of the current page to be parsed. In light of this, we collect a small dataset with this characteristic by crawling data from arXiv, which we will release to foster future research on VrDU models able to exploit the context. Note that Nougat, which we use as the base model, was trained on arXiv papers published on the website before January 2024. Thus, for our dataset, we select only papers released after January 2024 to avoid feeding the Encoder model with data already seen during its training. We download 6933 multi-page documents, and we perform the same preprocessing steps as in~\cite{blecher2023nougat}\footnote{\url{https://github.com/facebookresearch/nougat}}, resulting in 7724 annotated pages. In fact, due to the necessity of linking the visual parts of the images to the textual ground truth, after these preprocessing steps, only a portion of the image-markup pairs are deemed sufficiently high-quality and maintained in the dataset, with an acceptance rate of about 47\%. In our dataset, there can be four scenarios, depending on the availability of the previous and the next page as context for parsing the current page. These are summarized in~\Cref{tab:dataset_stats} and are as follows:
\begin{itemize}[topsep=0pt]
    \item Only the current page is available (the input to $\mu$gat will contain empty-page embeddings to replace the previous and the next page);
    \item The previous and the current pages are available, but not the next one (this latter is replaced with empty-page embeddings when preparing the input to $\mu$gat);
    \item The current and next pages are available, but not the previous one (this latter is replaced with empty-page embeddings when preparing the input to $\mu$gat);
    \item Both the previous and next page are available for the current page.
\end{itemize}
When composing our dataset, we make sure to maintain a balanced combination of these scenarios in order to facilitate the training of a more general adapter. Our dataset splits follow the 80-10-10 rule, obtaining 7690 samples for training, 962 for validation, and 913 for test. We provide detailed statistics in~\cref{tab:dataset_stats}.

\begin{table}[t]
    \centering
    \scriptsize
    \setlength{\tabcolsep}{2.5pt}
    \caption{arXiv dataset Statistics. For each split, we report the different types of context that can be provided to the model. Given the current page (curr), there can be the previous page (prev.) only, the next page (next) only, and both or neither available.}
    \label{tab:dataset_stats}
    \begin{tabular}{ccc c ccc}
       \toprule
       \textbf{prev.} & \textbf{curr.} & \textbf{next} && \textbf{Training} & \textbf{Validation} & \textbf{Test}\\
       \midrule
               & \cmark &         && 1276 & 156 & 165 \\
        \cmark & \cmark &         && 1758 & 207 & 210 \\
               & \cmark & \cmark  && 1758 & 208 & 210 \\
        \cmark & \cmark & \cmark  && 2898 & 391 & 328 \\
        \midrule
        \multicolumn{3}{c}{\textbf{total}} && 7690 & 962 & 913 \\
        \bottomrule       
    \end{tabular}
\end{table}

\tit{Tables}
In order to simulate the regesta use-case more closely, we build a synthetic dataset of three-column, full-page tables. Specifically, we select ten Pandoc\footnote{\url{https://pandoc.org/index.html}} templates\footnote{Selected among those in \url{https://github.com/cab-1729/Pandoc-Themes}} and fill them with text taken from parts of the papers in our arXiv dataset. In this way, the major challenge of the tables dataset will consist in the layout rather than in the text. The table documents are generated by paying attention to having a good variety of styles (in terms of fonts and table decorations) and table format. In this respect, we include both tables with columns of equal width, and more \textit{regesta-like} tables, in which the first two columns are smaller and contain fewer words than the third column. In fact, recall that in the regesta, the first two columns contain information about the date and place, while the third contains the summarized text of the referenced full-length document. In total, we obtain 399 samples that can be used for training or fine-tuning, 36 samples for validation, and 38 for test.

\tit{Regesta Pontificum Romanorum}
\label{sec:method:data:regesta}
In order to evaluate the performance of our models on the selected regesta use-case, we semi-automatically annotate 12 pages from the \textit{Regesta Pontificum Romanorum inde a 1198 ad 1304}. To this end, we feed GPT-4V with each page image and prompt it to parse the image into a \LaTeX~table that follows the three-column Regesta structure. Afterward, we manually correct both the transcription and formatting errors. Note that annotating the regesta is a time-consuming task, which also requires multidisciplinary expertise (\eg, certain proficiency both in markup and Latin languages). For this reason, we collect a small amount of pages. This set constitutes our evaluation benchmark for the trained models on the challenging regesta use-case. 
\section{Experiments}
\label{sec:experiments}

\subsection{Implementation and Training details}
We consider Nougat-Small~\cite{blecher2023nougat} as the baseline model, with Swin-Base as the Visual Encoder~\cite{liu2021swin} and a 4-layer mBART~\cite{lewis2019bart} Decoder. Following~\cite{blecher2023nougat}, we render the documents as images at a resolution of 96 DPI and resize and pad them to reach the resolution (896, 672) for compatibility with Swin-Base. This results in $P$=588 embedding vectors with $d=1024$. The proposed Adapter is composed of 2 layers with 4 learnable tokens. We train the overall $\mu$gat model for 50 epochs on the proposed academic papers dataset. Note that we keep the Encoder weights frozen, train the Adapter by using AdamW~\cite{loshchilov2017decoupled} as optimizer, with learning rate 5e-4, and fine-tune the Decoder with the same optimizer but learning rate 5e-5. Every epoch, the learning rates are reduced by a factor $\gamma$=0.9996 until they reach 2e-6, with an exponential scheduler. Note that we run an experiment in which we fine-tune on the synthetic tables. To this end, we use a learning rate of 5e-5 and fine-tune the models for 30 epochs.

\subsection{Experiment setup}
For validating our proposed approach, we consider the three datasets of documents described in~\cref{sec:method_data}. Recall that these are academic papers (which we refer to as \textbf{arXiv}), rendered tables (which we refer to as \textbf{tables}), and pages from the Regesta Pontificum Romanorum (which we refer to as \textbf{RPR}). 

As for the evaluation scores, in line with Document Parsing literature, we report the performance in terms of the following metrics: Edit Distance (\textbf{ED})~\cite{levenshtein1966binary}, which measures the number of character substitutions, insertions, and deletions to transform one string into another; BLEU (\textbf{B})~\cite{papineni2002bleu}, which measures the number of matching n-grams between a source and target text; METEOR (\textbf{M})~\cite{banerjee2005meteor}, which considers also the recall of the target n-grams; Precision (\textbf{P}), and Recall (\textbf{R}). Note that these two latter scores are obtained by collecting the words in the ground truth and in the model output prediction into two sets and calculating the number of words in the prediction set that also appear in the ground truth (for the precision) and the number of words in the ground truth set that also appear in the prediction (for the recall).

\subsection{Results}

\begin{table}[t]
    \centering
    \scriptsize
    \setlength{\tabcolsep}{2.5pt}
    \caption{Comparison between \textit{\textmu}gat, Nougat-S (our base model) applied in direct transfer and fine-tuned on the same data used to train our model (Nougat-S$^*$), and Nougat-B applied in direct transfer.}\label{tab:general-results}
    \begin{tabular}{l c ccccc c ccccc}
    \toprule
    && \multicolumn{5}{c}{\textbf{arXiv}} && \multicolumn{5}{c}{\textbf{RPR}}                        \\
    \cmidrule{3-7} \cmidrule{9-13}
    && \textbf{ED}&\textbf{B}&\textbf{M}&\textbf{P}&\textbf{R} && \textbf{ED}&\textbf{B}&\textbf{M}&\textbf{P}&\textbf{R} \\
    \midrule
    \textbf{Nougat-S}     && 15.6 & 80.1 & 86.3 & 90.4 & 88.7 && 75.4 & 3.2 & 11.0 & 36.4 & 15.0 \\
    \textbf{Nougat-S*} && 10.4 & 84.7 & 89.1 & 91.9 & 89.5 && 75.4 & 3.2 & 11.0 & 36.4 & 16.0 \\
    \textbf{Nougat-B}     && 15.6 & 80.1 & 86.3 & 90.5 & 88.7 && 75.4 & 3.2 & 11.0 & 36.4 & 16.0 \\
    \textbf{\textit{\textmu}gat}     && ~8.8 & 87.1 & 91.4 & 92.2 & 91.2 && 72.1 & 4.8 & 11.8 & 18.9 & 14.3 \\
    \bottomrule
    \end{tabular}
\end{table}

\tit{Quantitative Comparison} 
First, we perform a quantitative comparison against the baseline Nougat-Small (which we refer to as Nougat-S) model, both directly applied to the arXiv and RPR samples and fine-tuned on the training split of arXiv (we refer to this version as Nougat-S*), and the larger Nougat-Base model (which we refer to as Nougat-B). We show the results on arXiv and RPR in~\Cref{tab:general-results}. As we can see, our model performs better than the competitors. Another aspect that emerges from the results is that the RPR dataset, to which we directly apply all the compared models without any fine-tuning, is still very challenging, even for our model.

\tit{Role of the context}
Moreover, in~\Cref{tab:context-results}, we show the validation results for our $\mu$gat model and the Nougat-S* baseline on the arXiv test set further divided depending on the context available for each page to be parsed. As we can see, while the performance of the compared models is very close when the isolated current page is provided, the performance of $\mu$gat consistently increases as more context is provided, indicating that our strategy allows exploiting inter-page structure for parsing. 

\begin{table}[t]
    \centering
    \scriptsize
    \setlength{\tabcolsep}{1.3pt}
    \caption{Performance of Nougat-S* and $\mu$gat on pages of the arXiv test set for which it is available just the current page, also the previous or the next, and both.}\label{tab:L_N_ablation}
    \begin{tabular}{c c ccc c ccccc}
    \toprule
    && \textbf{prev.} & \textbf{curr.}  & \textbf{next} && \textbf{ED}&\textbf{B}&\textbf{M}&\textbf{P}&\textbf{R}\\
    \midrule
    \multirow{4}{*}{\textbf{Nougat-S*}} 
    &&        & \cmark &         && 13.4& 80.2& 85.6& {88.0}& 86.3 \\
    && \cmark & \cmark &         && 11.2& 83.9& 88.7& 91.7& 89.0 \\
    &&        & \cmark & \cmark  && 10.9& 84.6& 88.8& 92.5& 89.0 \\
    && \cmark & \cmark & \cmark  && ~7.6& 88.1& 91.6& 93.7& 91.7 \\
    \midrule
    \multirow{4}{*}{\textbf{\textit{\textmu}gat}}  
    &&        & \cmark &         && 12.3& 81.8& 87.4& {88.0}& 87.5 \\
    && \cmark & \cmark &         && ~9.3& 86.7& 91.4& 92.5& 91.2 \\
    &&        & \cmark & \cmark  && ~9.1& 87.5& 91.3& 92.7& 91.2 \\
    && \cmark & \cmark & \cmark  && ~6.5& 89.8& 93.4& 93.8& 93.2 \\
    \bottomrule
    \end{tabular}
    \label{tab:context-results}
\end{table}

\tit{Role of the training or fine-tuning the dataset}
To evaluate the role of the training and fine-tuning, we perform an analysis in which we fine-tune the baseline Nougat-S on the arXit training set used for training $\mu$gat and further fine-tune both models on the tables dataset. From the results reported in~\Cref{tab:ft_ablation}, we observe that fine-tuning improves the performance of both models but does not help much when directly applying them to the RPR pages. It is worth noting that Nougat-S benefits less than $\mu$gat from fine-tuning, especially on tables. 

\begin{table}[t]
    \centering
    \scriptsize
    \setlength{\tabcolsep}{1.3pt}
    \caption{Ablation on the dataset used for fine-tuning (ft) or training (tr) the Adapter of \textit{\textmu}gat and decoder of Nougat-S (our base model) or \textit{\textmu}gat.}\label{tab:ft_ablation}
    \begin{tabular}{cc c ccccc c ccccc c ccccc}
    \toprule
    & && \multicolumn{5}{c}{\textbf{arXiv}} && \multicolumn{5}{c}{\textbf{tables}} && \multicolumn{5}{c}{\textbf{RPR}} \\
    \cmidrule{4-8} \cmidrule{10-14} \cmidrule{16-20}
    & \textbf{tr/ft}   && \textbf{ED}&\textbf{B}&\textbf{M}&\textbf{P}&\textbf{R} && \textbf{ED}&\textbf{B}&\textbf{M}&\textbf{P}&\textbf{R} && \textbf{ED}&\textbf{B}&\textbf{M}&\textbf{P}&\textbf{R} \\
    \midrule
    \multirow{3}{*}{\textbf{Nougat-S}}  & -              && 15.6 & 80.1 & 86.3 & 90.4 & 88.7 && 63.3 & 31.1 & 59.5 & 79.7 & 65.9 && 75.4 & 3.2 & 11.0 & {36.4} & {15.0}\\
    & arXiv          && 10.4 & 84.7 & 89.1 & 91.9 & 89.5 && 63.7 & 27.6 & 41.3 & 69.6 & 44.6 && 75.4 & 3.2 & 11.0 & {36.4} & {15.0}\\
    & arXiv, tables  && 19.7 & 75.2 & 82.7 & 84.6 & 83.8 && 58.7 & 35.7 & 57.7 & 73.8 & 62.4 && 75.4 & 3.2 & 11.0 & {36.4} & {15.0}\\
    \midrule
    \multirow{2}{*}{\textbf{\textit{\textmu}gat}}  & arXiv          && {~8.8} & {87.1} & {91.4} & {92.2} & {91.2} && 61.4 & 32.7 & 51.1 & 68.1 & 55.3 && {72.1} & {4.8} & {11.8} & 18.9 & 14.3\\
    & arXiv, tables  && 18.5 & 76.4 & 82.2 & 84.2 & 82.6 && {~8.6} & {85.5} & {94.2} & {96.4} & {94.5} && 84.9 & 0.7 & 04.8 & 23.3 & ~7.1 \\
    \bottomrule
    \end{tabular}
\end{table}

\tit{Ablation on the number of Adapter layers and tokens}
In~\Cref{tab:L_N_ablation}, we show an ablation analysis by varying the number of Adapter layers $L$ and of the learnable tokens $N$. We notice that increasing $L$ beyond 2 does not improve the performance, possibly also due to the small size of the training dataset. As for the number of tokens, we empirically find that 4 is a good trade-off between representation capabilities and generalization.

\begin{table}[t]
    \centering
    \scriptsize
    \setlength{\tabcolsep}{1.6pt}
    \caption{Ablation on the number of Adapter layers and tokens in our \textit{\textmu}gat model.}\label{tab:L_N_ablation}
    \begin{tabular}{cc c ccccc c ccccc c ccccc}
    \toprule
    & && \multicolumn{5}{c}{\textbf{arXiv}} && \multicolumn{5}{c}{\textbf{tables}} && \multicolumn{5}{c}{\textbf{RPR}} \\
    \cmidrule{4-8} \cmidrule{10-14} \cmidrule{16-20}
    \textbf{\#layers}   & \textbf{\#tokens}   && \textbf{ED}&\textbf{B}&\textbf{M}&\textbf{P}&\textbf{R} && \textbf{ED}&\textbf{B}&\textbf{M}&\textbf{P}&\textbf{R} && \textbf{ED}&\textbf{B}&\textbf{M}&\textbf{P}&\textbf{R} \\
    \midrule
    \multirow{3}{*}{2} & 2  && 10.2 & 84.9 & 89.7 & 91.4 & 89.8 && 67.3 & 24.4 & 39.2 & 69.4 & 43.1 && 84.8 & 1.2 & ~4.1 & 22.7 & ~6.2 \\
    & 4      && ~8.8 & 87.1 & 91.4 & {92.2} & {91.2} && {61.4} & {32.7} & {51.1} & 68.1 & {55.3} && {72.1} & {4.8} & {11.8} & 18.9 & {14.3} \\
    & 8      && ~9.6 & 86.1 & 90.7 & 91.7 & 90.7 && 68.5 & 20.4 & 35.0 & 58.6 & 38.6 && 76.7 & 2.4 & ~6.4 & 16.6 & ~8.4 \\
    \midrule
    \multirow{3}{*}{4} & 2      && ~9.4 & 86.4 & 90.7 & 91.9 & 90.7 && 64.1 & 27.5 & 46.2 & 71.8 & 50.4 && 77.3 & 2.6 & ~7.1 & 17.1 & ~9.6 \\
    & 4     && ~9.9 & 85.7 & 90.4 & 91.6 & 90.5 && 65.4 & 27.3 & 40.9 & {75.6} & 43.6 && 78.0 & 2.6 & ~8.3 & {24.1} & 11.3 \\
    & 8      && ~9.6 & 86.0 & 90.6 & 91.5 & 90.6 && 63.4 & 31.9 & 47.4 & 73.1 & 51.7 && 88.3 & 0.8 & ~2.5 & 14.4 & ~3.5 \\
    \bottomrule
    \end{tabular}
\end{table}

\begin{figure}[t]
    \centering
    \scriptsize
    \setlength{\tabcolsep}{1em}
    \begin{tabular}{m{.6em}c  c }
        \midrule
        \rotatebox[origin=l]{90}{\textbf{Original}} &
         \makecell{\includegraphics[height=4.25cm]{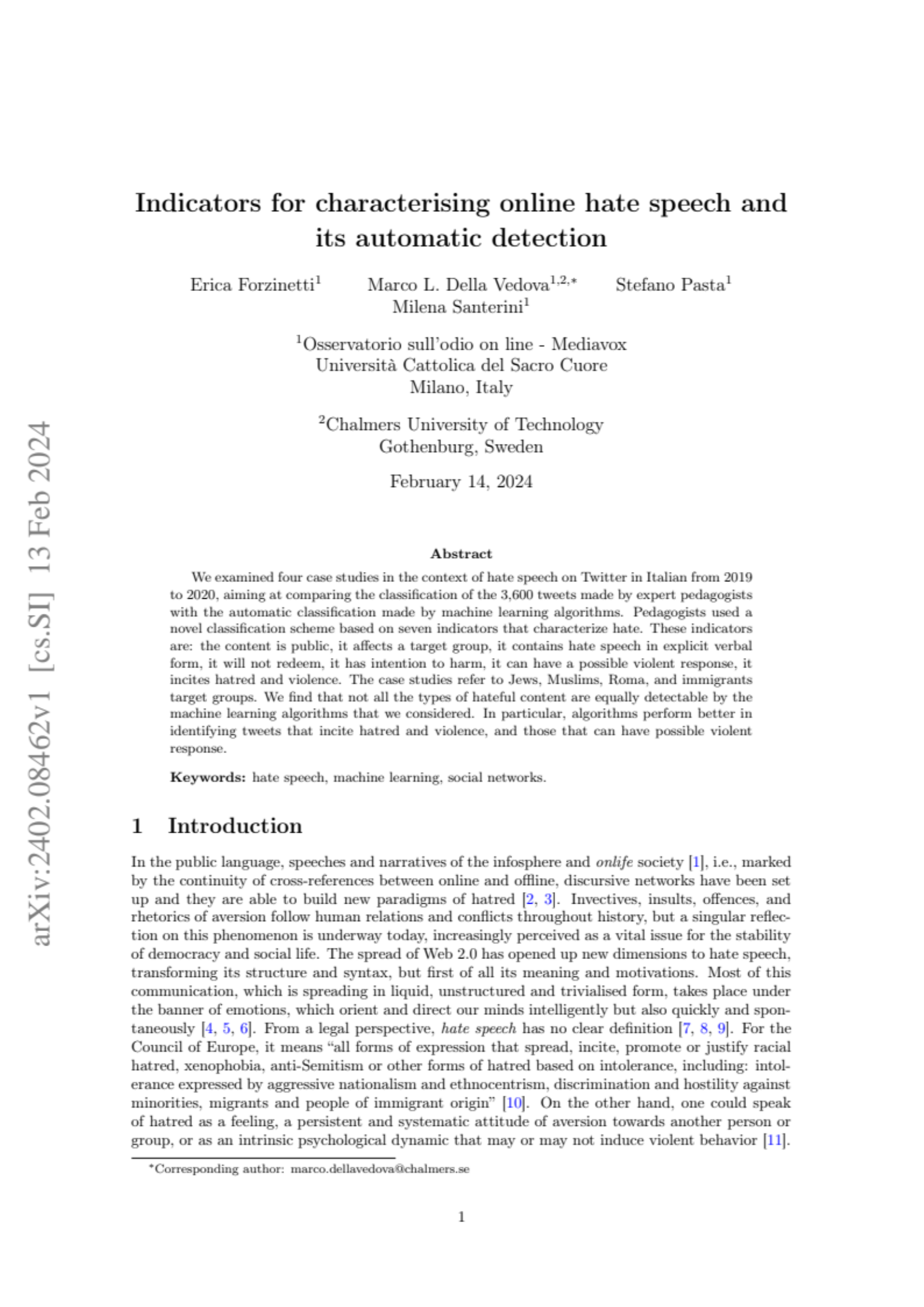}} & 
         \makecell{\includegraphics[height=4.25cm]{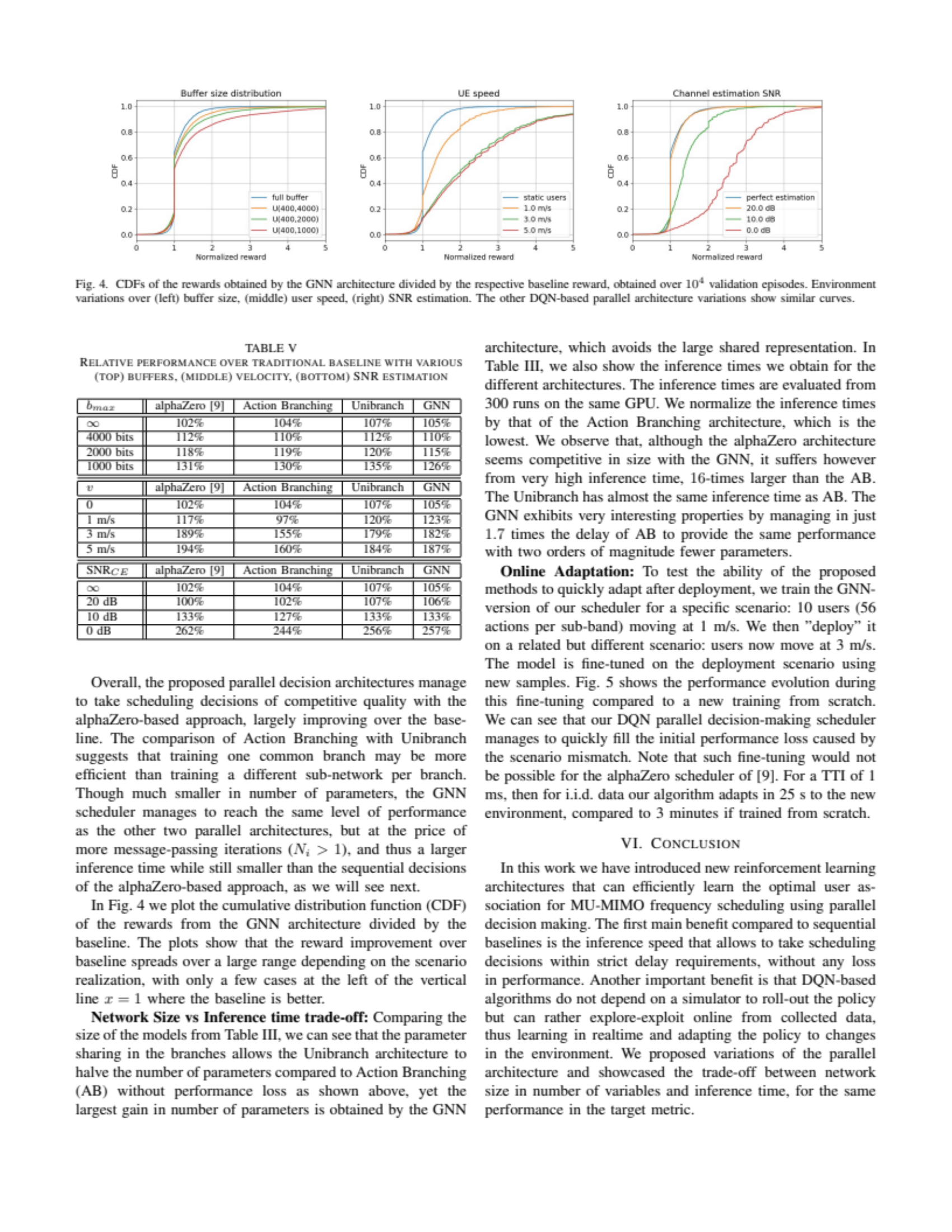}} \\
         \midrule
        \rotatebox[origin=l]{90}{\textbf{Ground Truth}} &  
         \makecell{\includegraphics[height=4.25cm]{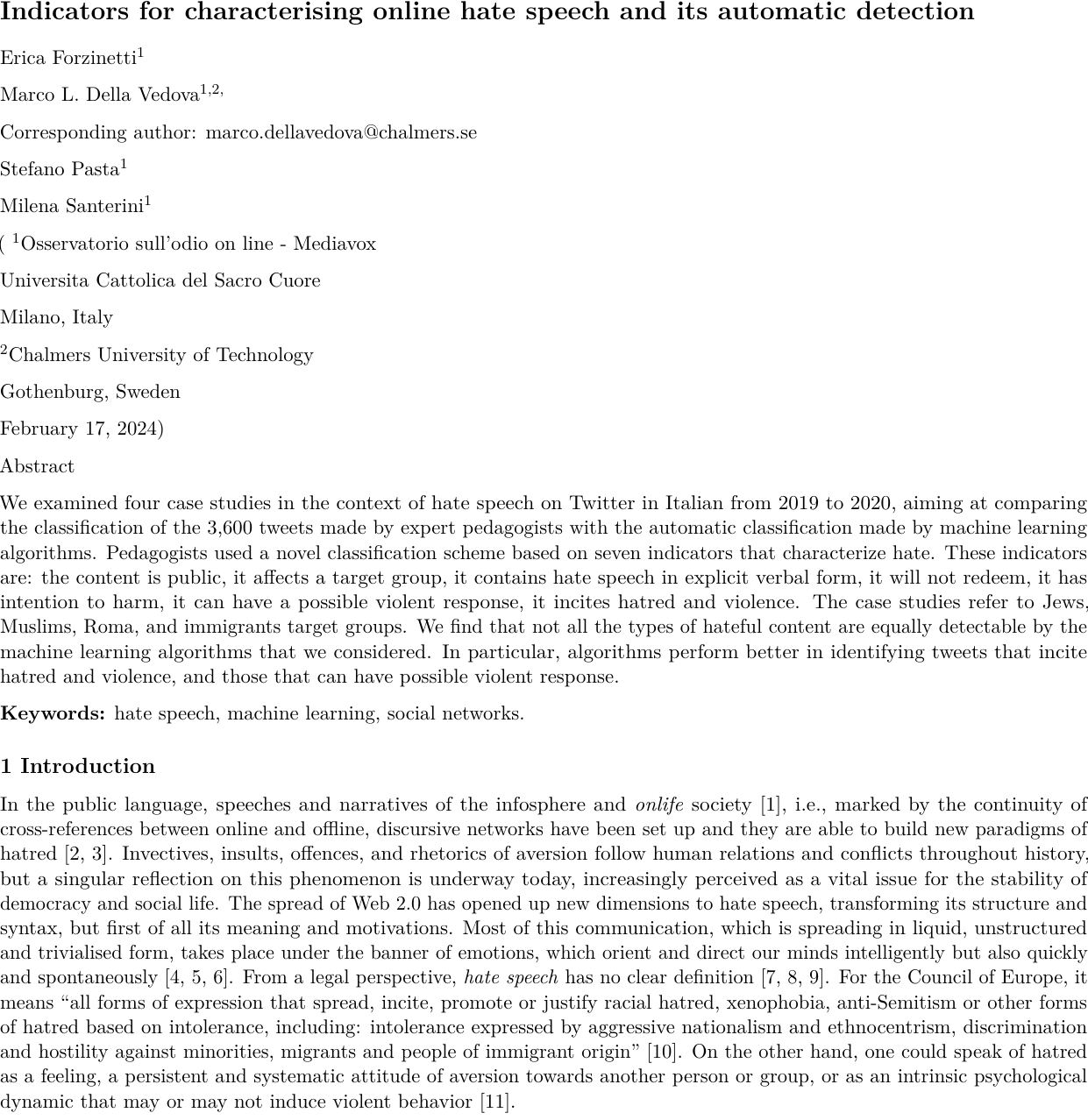}} & 
         \makecell{\includegraphics[height=4.25cm]{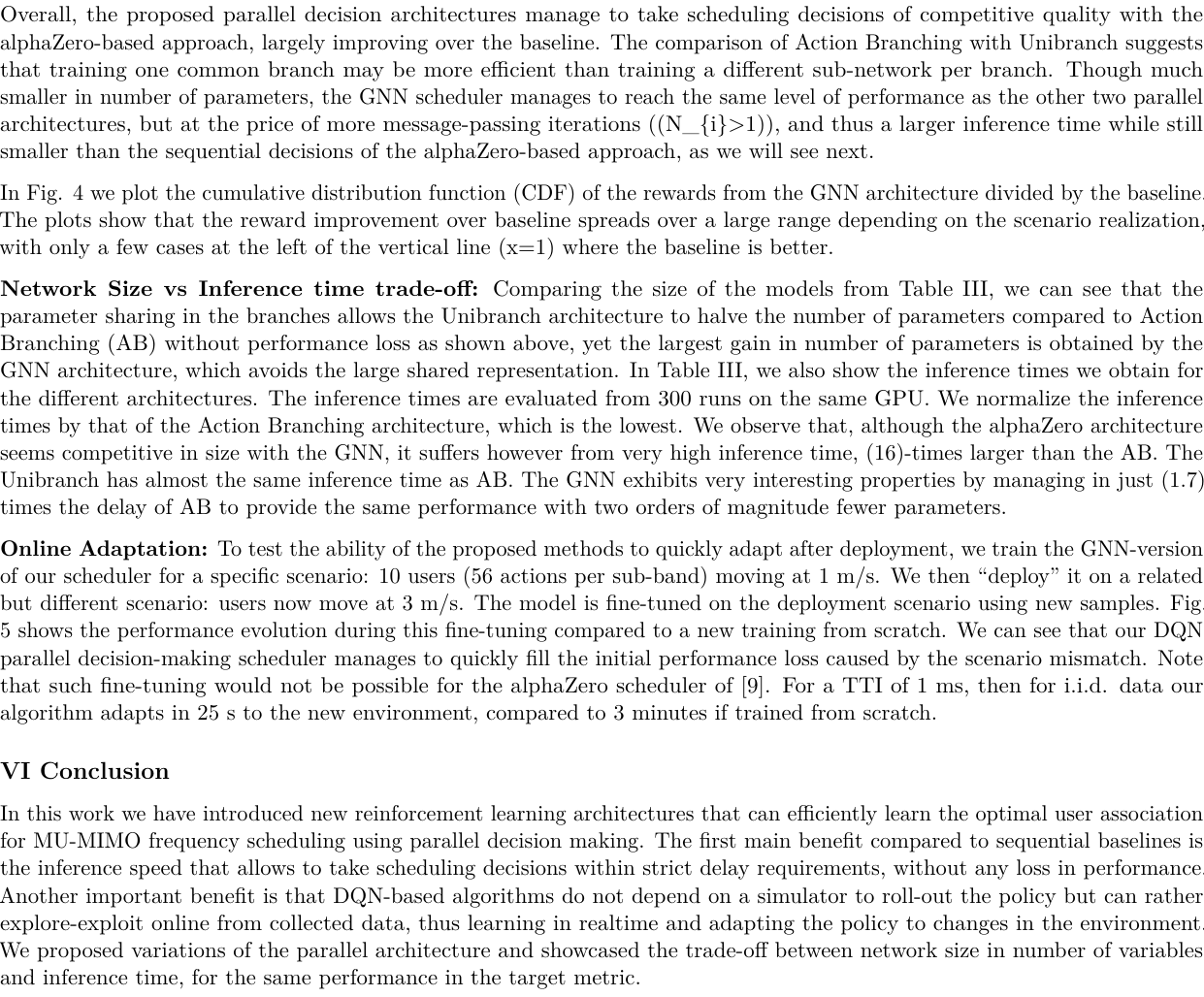}} \\
         \midrule
        \rotatebox[origin=l]{90}{\textbf{Nougat-S*}} &
         \makecell{\includegraphics[height=4.25cm]{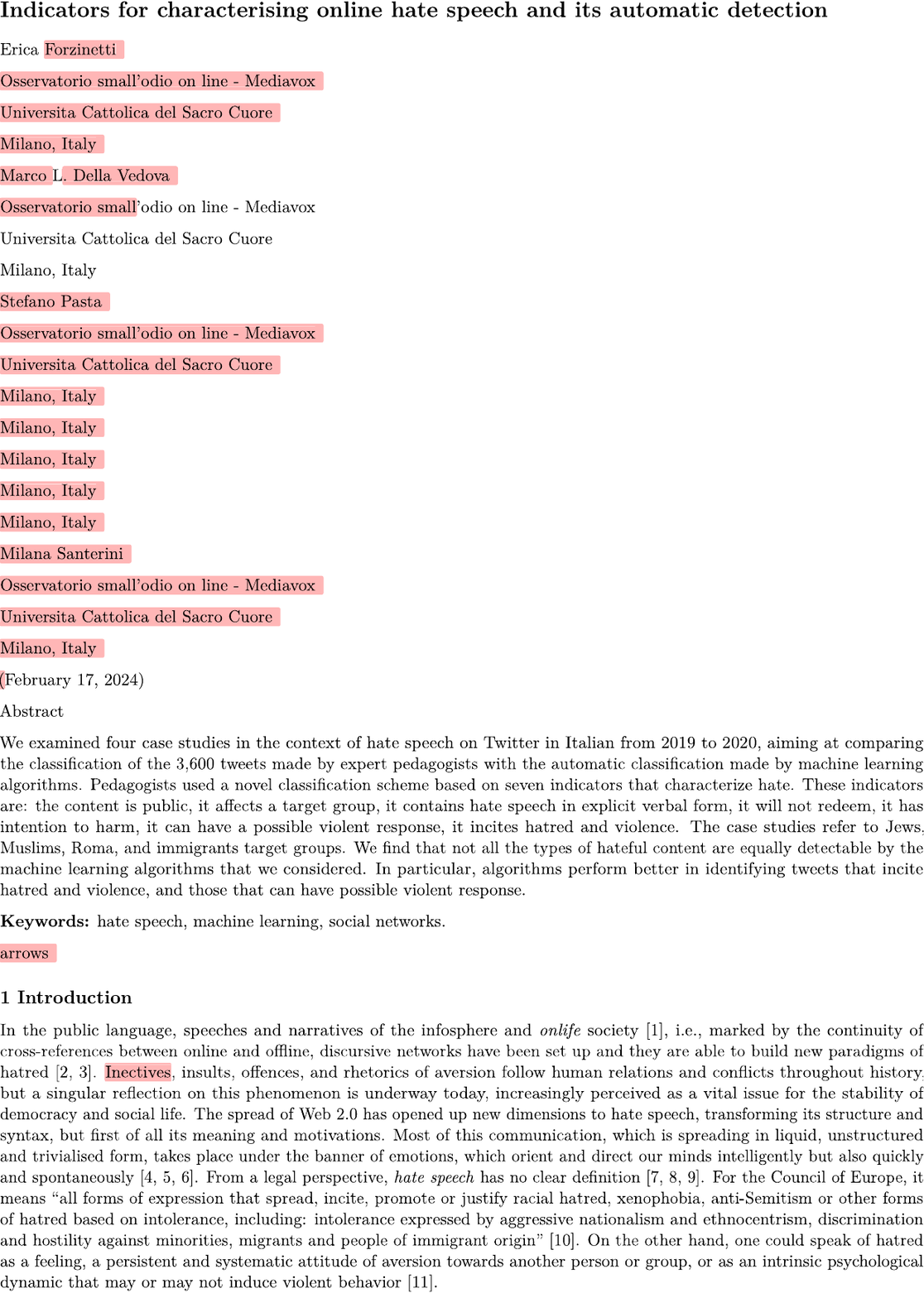}} & 
         \makecell{\includegraphics[height=4.25cm]{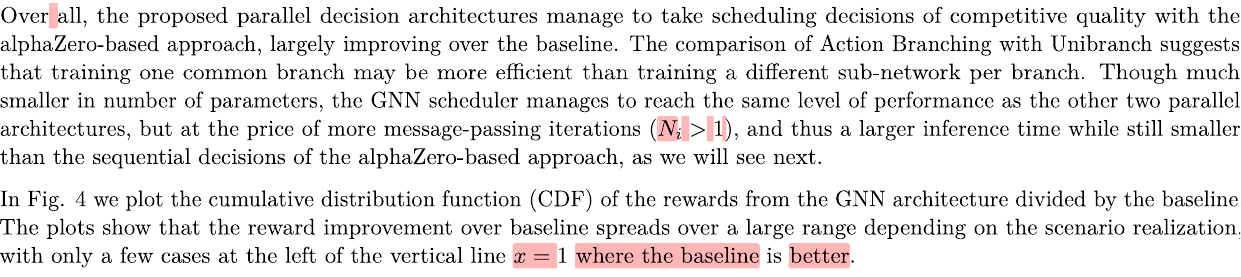}} \\
         \midrule
        \rotatebox[origin=l]{90}{\textbf{\textit{\textmu}gat}} &
         \makecell{\includegraphics[height=4.25cm]{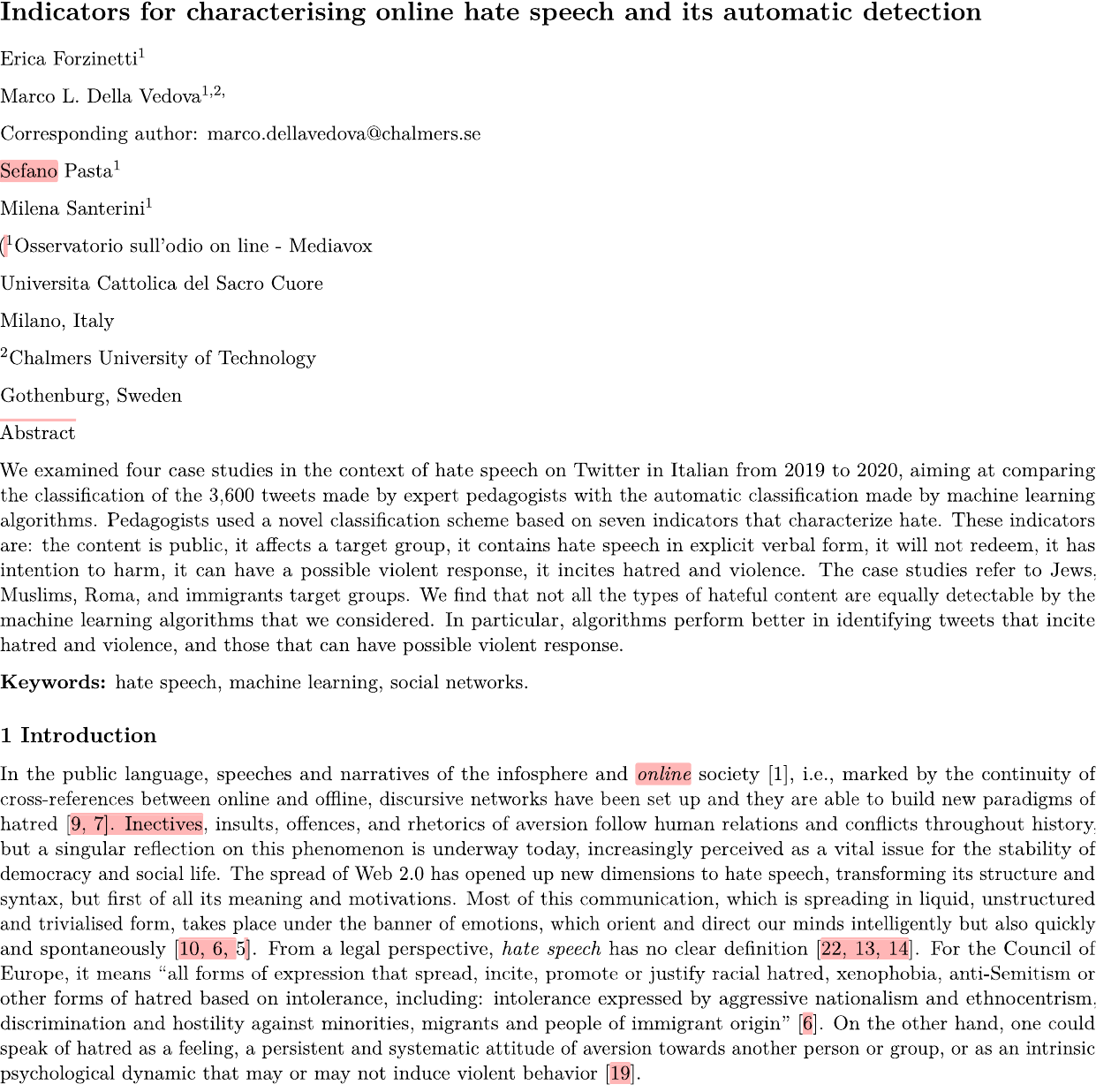}} & 
         \makecell{\includegraphics[height=4.25cm]{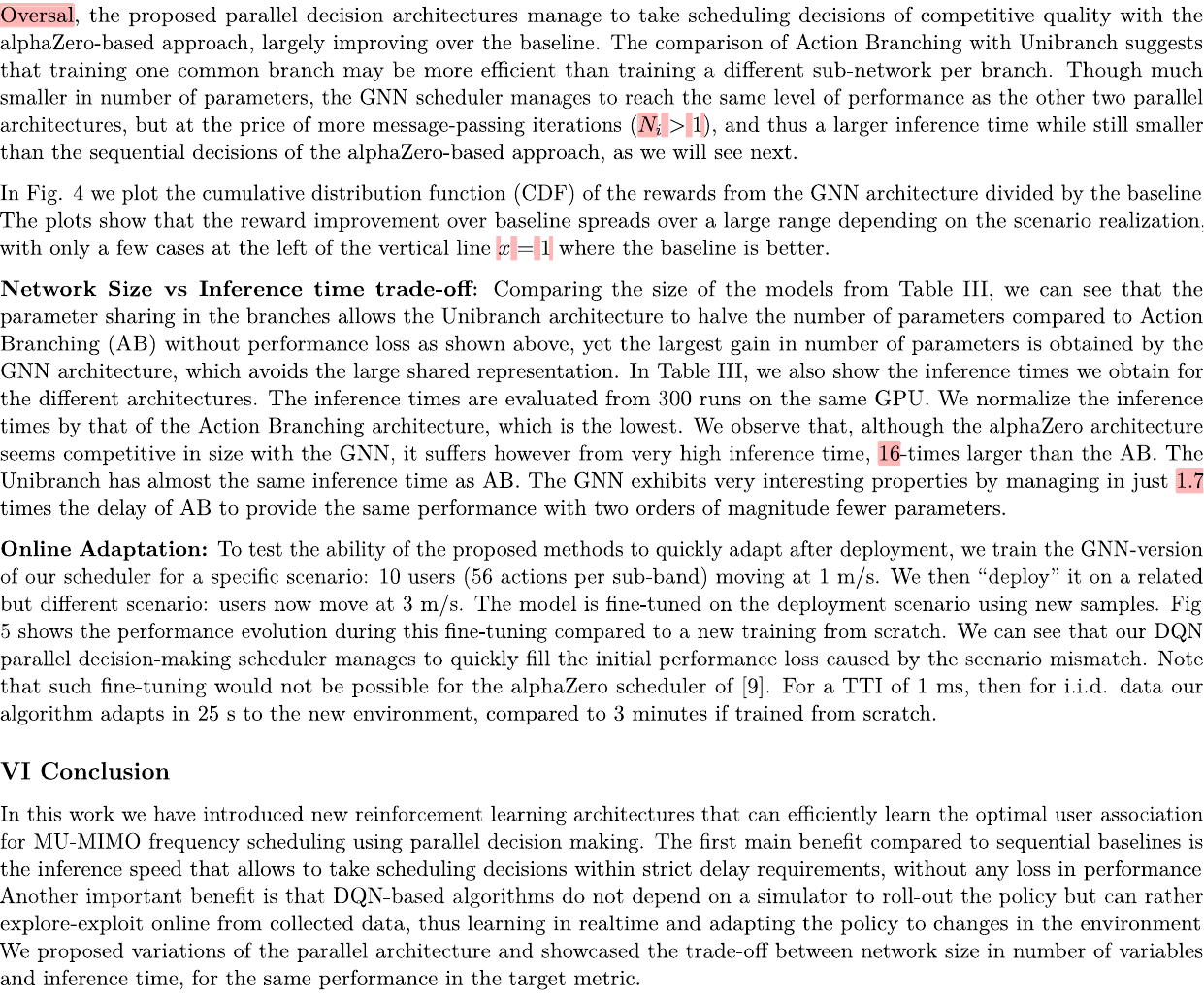}} \\ 
         \midrule
    \end{tabular}
    \caption{Qualitative results on two pages from arXiv, when no context is available (left) and full context is available (right). We report the Ground Truth and highlight the errors made by $\mu$gat and Nougat-S* in their \LaTeX-rendered output.}
    \label{fig:qualitatives_arXiv}
\end{figure}

\begin{figure}[t]
    \centering
    \scriptsize
    \setlength{\tabcolsep}{1em}
    \begin{tabular}{m{.6em}c c }
    \midrule
        \rotatebox[origin=l]{90}{\textbf{Original}} &
         \makecell{\includegraphics[height=4.25cm]{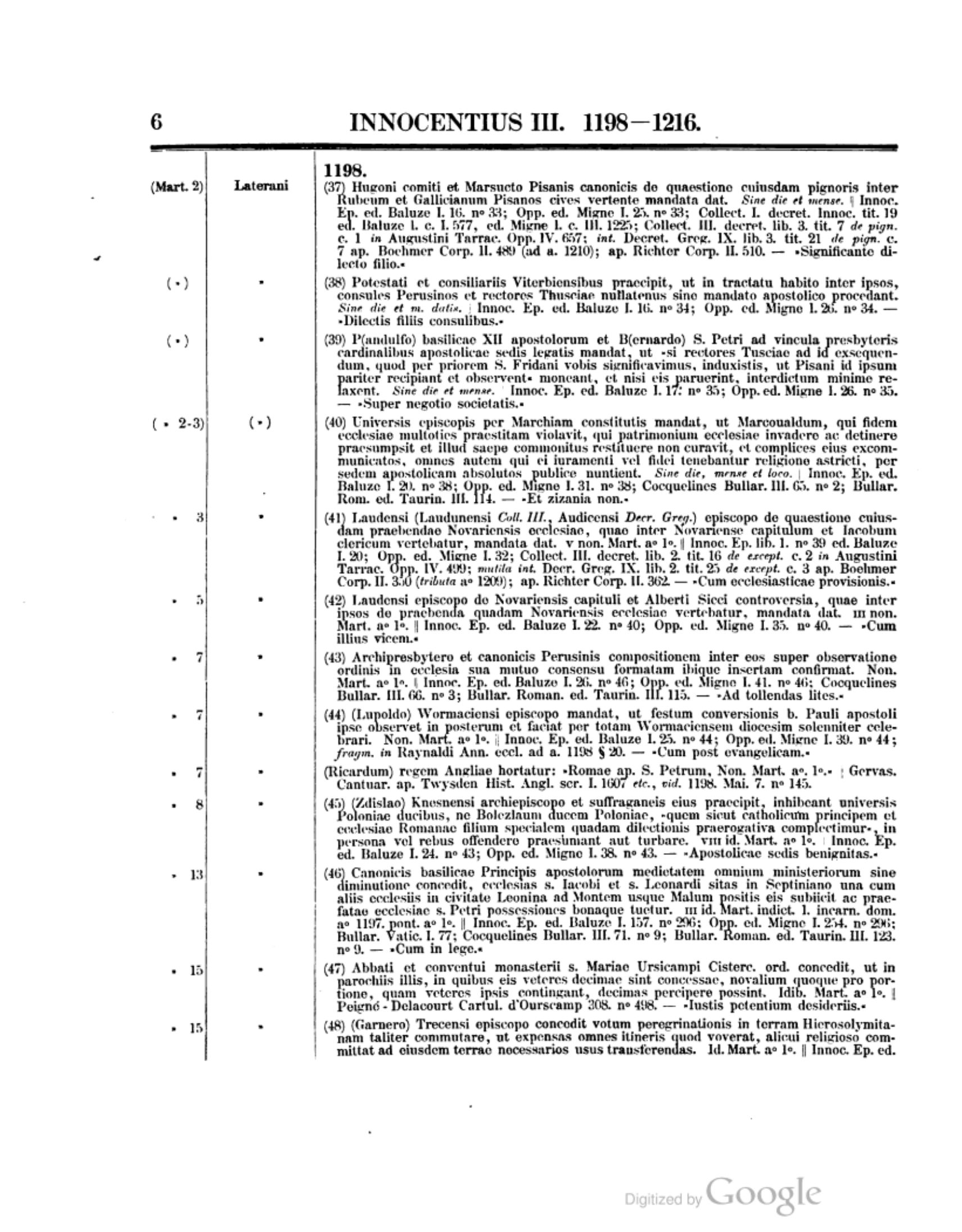}} & 
         \makecell{\includegraphics[height=4.25cm]{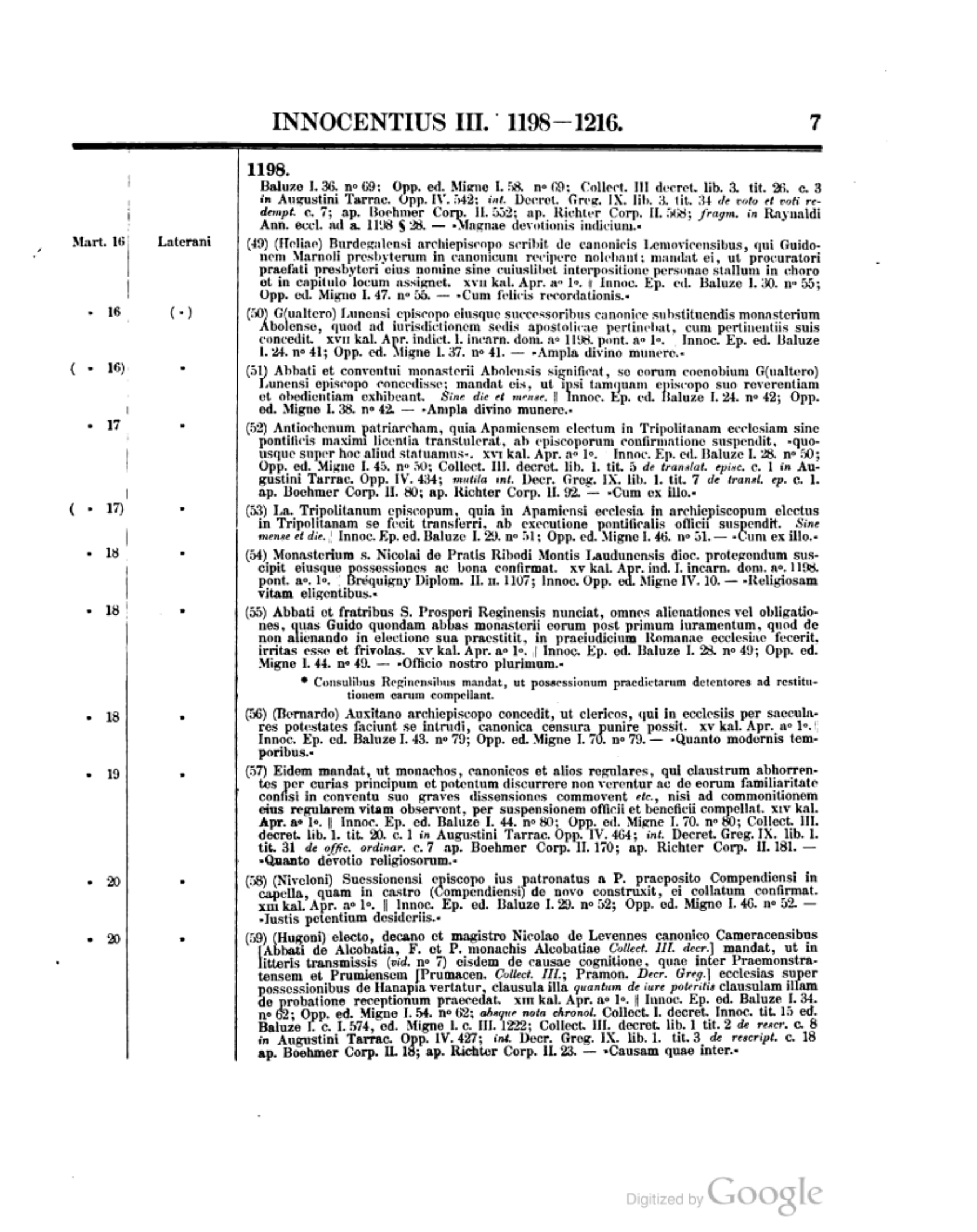}} \\
    \midrule
        \rotatebox[origin=l]{90}{\textbf{Ground Truth}} &  
         \makecell{\includegraphics[height=4.25cm]{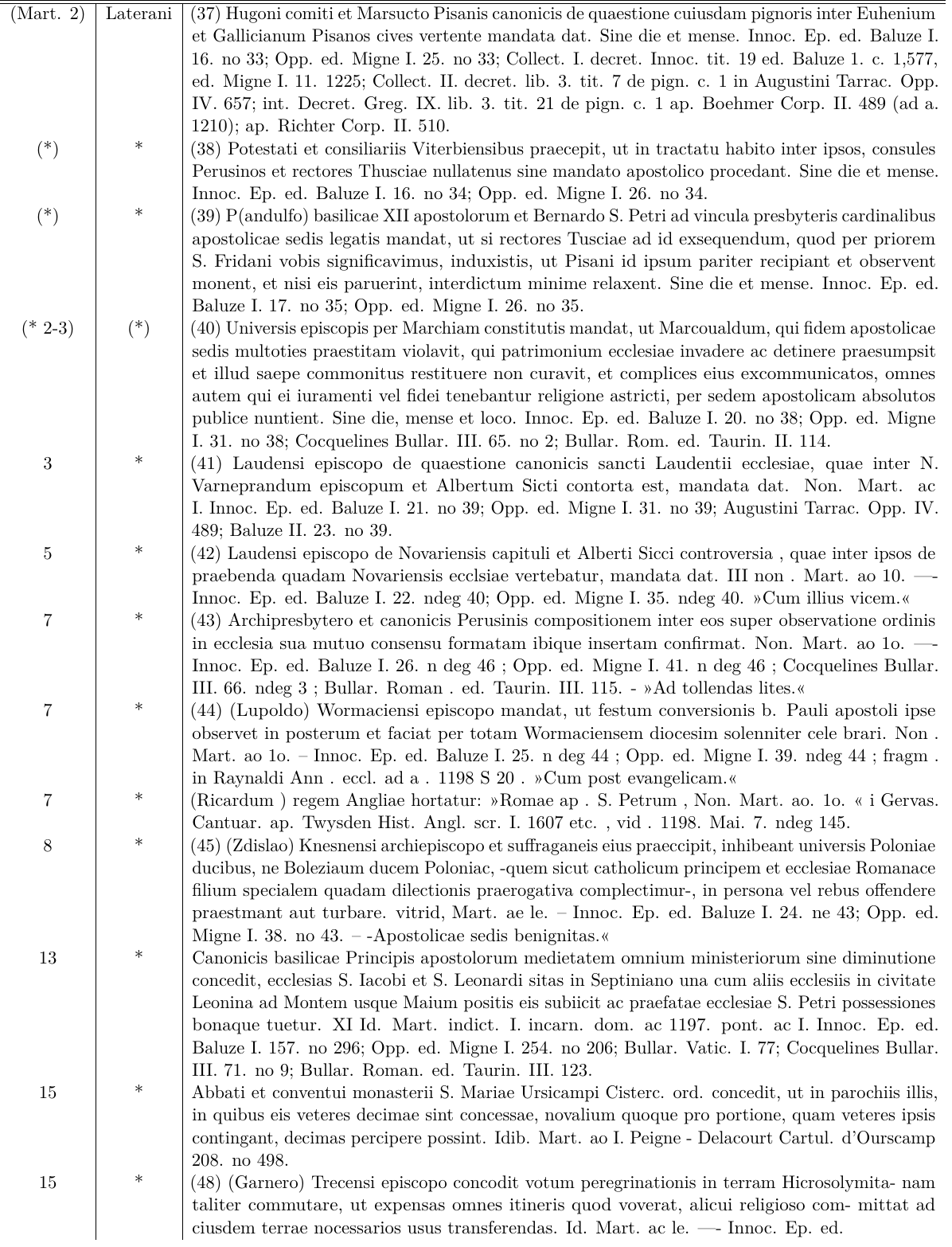}} & 
         \makecell{\includegraphics[height=4.25cm]{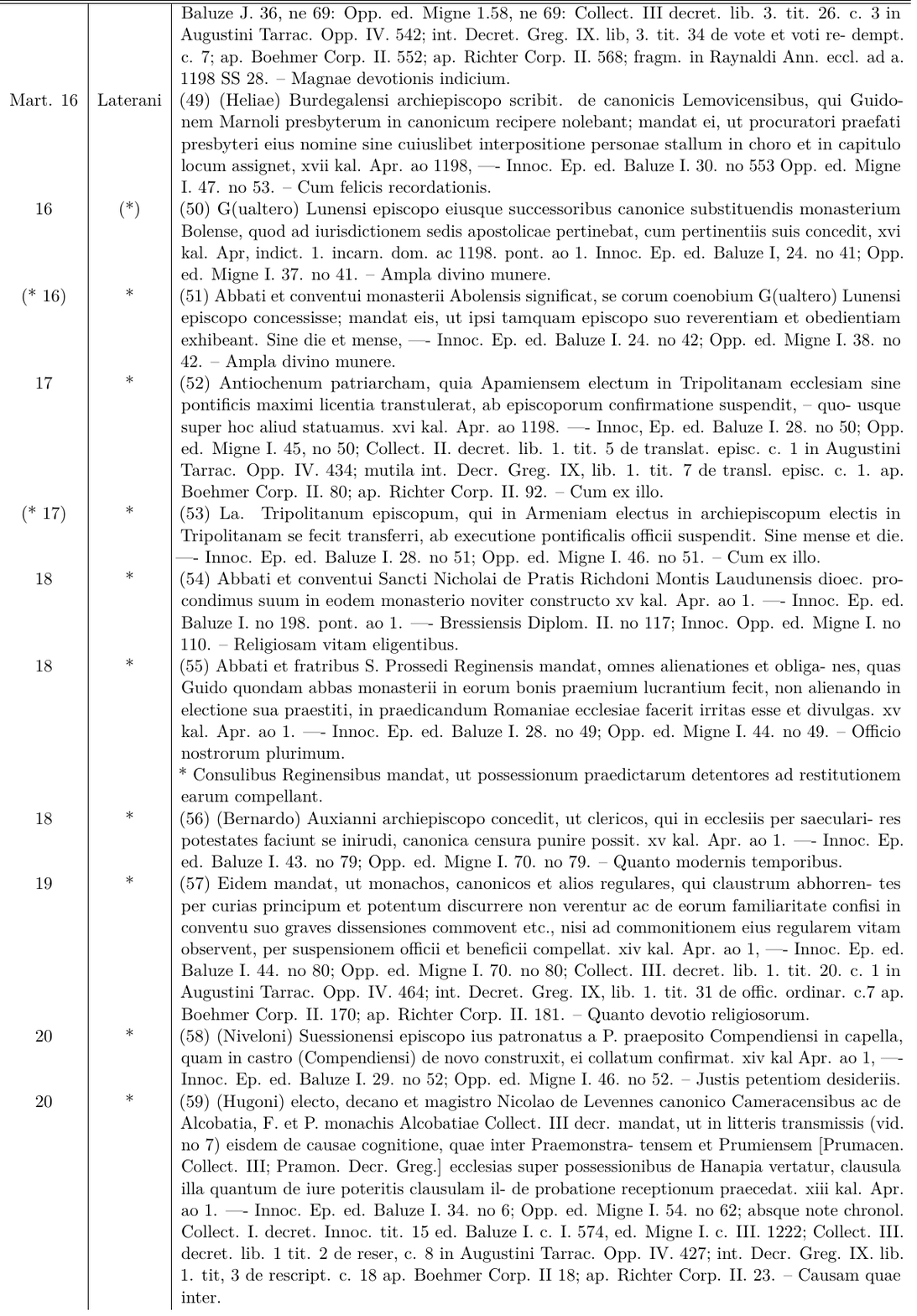}} \\
    \midrule
        \rotatebox[origin=l]{90}{\textbf{\textit{\textmu}gat}} &
         \makecell{\includegraphics[height=4.25cm]{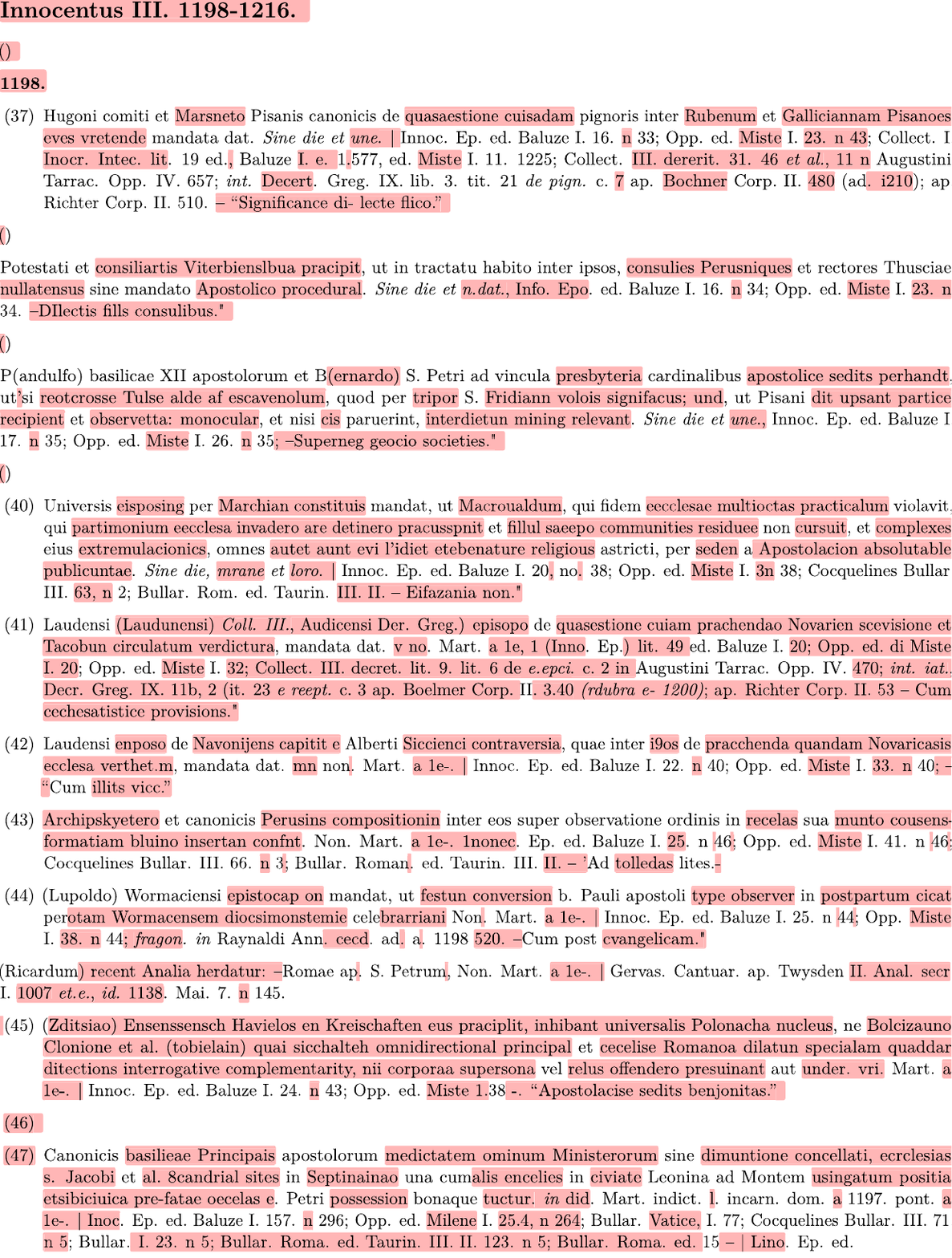}} & 
         \makecell{\includegraphics[height=4.25cm]{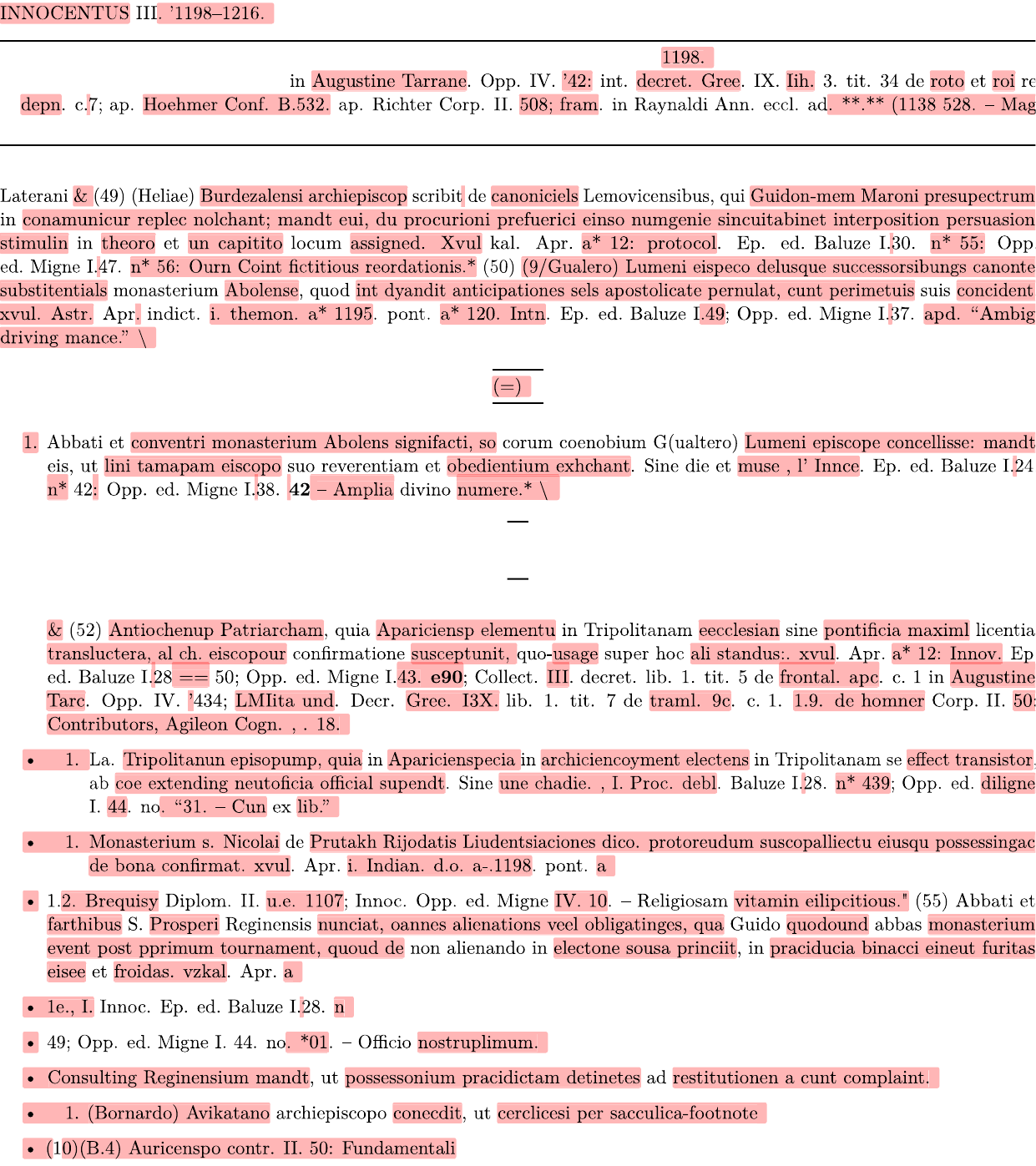}} \\ 
    \midrule
    \end{tabular}
    \caption{Qualitative results on two pages from the challenging RPR. We report the Ground Truth and highlight the errors made by $\mu$gat in its \LaTeX-rendered output.}
    \label{fig:qualitatives_regesta}
\end{figure}

\tit{Qualitative results}
Finally, we repost a qualitative analysis of the performance of Nougat-S* and $\mu$gat on the arXiv dataset in~\cref{fig:qualitatives_arXiv} and of $\mu$gat on the RPR pages in~\cref{fig:qualitatives_arXiv}. The images for the ground truth and the models parsing output are obtained by rendering the markdown/\LaTeX in PDF by using Pandoc.
From~\cref{fig:qualitatives_arXiv}, we can observe that the output of $\mu$gat for the arXiv data is very close to the ground truth, especially when full context is provided. On the other hand, Nougat-S* misses some more details (like superscript numbers signaling footnotes) and even collapses into repeating the phrase \textit{Milano, Italy} a few times. 
As for RPR pages (\cref{fig:qualitatives_regesta}), we observe that our model struggles to capture the complex table-like structure but can extract most of the Latin text from the page, with some mistakes. We argue that this is a promising result since it shows that, even though the model was trained almost completely on English text, it exhibits some capability to also extract text in other languages and from low visual quality pages. This ability and the capability to deal with the complex layout of regesta could be further improved by fine-tuning the model on a larger amount of specific data, which we leave for future research.

\section{Conclusion}
\label{sec:conclusion}

In this work, we have investigated the limitations of current single-page parsing models in handling multi-page documents. This is critical in many real-world scenarios, particularly for documents where inter-page dependencies are crucial, such as regesta collections, which are a valuable tool for historians and humanities scholars.
To tackle this limitation, we have proposed a novel approach for context-enhanced document parsing, dubbed $\mu$gat. 
Our key contribution is the introduction of a novel adaptation strategy for pre-trained, OCR-free single-page document parsing models (specifically, the recently proposed Nougat), which extends these models for effectively utilizing the context provided by surrounding pages, resulting in significantly improved parsing consistency and accuracy.  

We have validated the efficacy of our approach on a specifically collected dataset of scientific papers consisting of pages with their context, rendered long tables, and pages from the Regesta Pontificum Romanorum. The results obtained demonstrate that leveraging information from surrounding pages can significantly improve the parsing accuracy of single-page models.
These encouraging results can contribute to the development of VrDU solutions able to deal with multi-page, intricately-structured documents, such as ancient documents and regesta, which are particularly relevant to Digital Humanities.

\section*{Acknowledgement}
This work was supported by the ``AI for Digital Humanities'' project funded by ``Fondazione di Modena'' and the PNRR project Italian Strengthening of ESFRI RI Resilience (ITSERR) funded by the European Union – NextGenerationEU.

\clearpage
\bibliographystyle{splncs04}
\bibliography{main}

\begin{thebibliography}{10}
\providecommand{\url}[1]{\texttt{#1}}
\providecommand{\urlprefix}{URL }
\providecommand{\doi}[1]{https://doi.org/#1}

\bibitem{appalaraju2021docformer}
Appalaraju, S., Jasani, B., Kota, B.U., Xie, Y., Manmatha, R.: {Docformer: End-to-end transformer for document understanding}. In: ICCV (2021)

\bibitem{appalaraju2024docformerv2}
Appalaraju, S., Tang, P., Dong, Q., Sankaran, N., Zhou, Y., Manmatha, R.: {Docformerv2: Local features for document understanding}. In: AAAI (2024)

\bibitem{banerjee2005meteor}
Banerjee, S., Lavie, A.: {METEOR: An automatic metric for MT evaluation with improved correlation with human judgments}. In: Proceedings of the acl workshop on intrinsic and extrinsic evaluation measures for machine translation and/or summarization. pp. 65--72 (2005)

\bibitem{biten2022latr}
Biten, A.F., Litman, R., Xie, Y., Appalaraju, S., Manmatha, R.: {Latr: Layout-aware transformer for scene-text vqa}. In: CVPR (2022)

\bibitem{blau2024gram}
Blau, T., Fogel, S., Ronen, R., Golts, A., Ganz, R., Ben~Avraham, E., Aberdam, A., Tsiper, S., Litman, R.: {GRAM: Global Reasoning for Multi-Page VQA}. In: CVPR (2024)

\bibitem{blecher2023nougat}
Blecher, L., Cucurull, G., Scialom, T., Stojnic, R.: {Nougat: Neural optical understanding for academic documents}. In: ICLR (2024)

\bibitem{davis2022end}
Davis, B., Morse, B., Price, B., Tensmeyer, C., Wigington, C., Morariu, V.: {End-to-end document recognition and understanding with dessurt}. In: ECCV (2022)

\bibitem{devlin2018bert}
Devlin, J., Chang, M.W., Lee, K., Toutanova, K.: {Bert: Pre-training of deep bidirectional transformers for language understanding}. In: NAACL (2018)

\bibitem{harley2015evaluation}
Harley, A.W., Ufkes, A., Derpanis, K.G.: {Evaluation of deep convolutional nets for document image classification and retrieval}. In: ICDAR (2015)

\bibitem{huang2022layoutlmv3}
Huang, Y., Lv, T., Cui, L., Lu, Y., Wei, F.: {Layoutlmv3: Pre-training for document ai with unified text and image masking}. In: ACM Multimedia (2022)

\bibitem{jaegle2021perceiver}
Jaegle, A., Gimeno, F., Brock, A., Vinyals, O., Zisserman, A., Carreira, J.: {Perceiver: General Perception with Iterative Attention}. In: ICML (2021)

\bibitem{jaume2019funsd}
Jaume, G., Ekenel, H.K., Thiran, J.P.: {Funsd: A dataset for form understanding in noisy scanned documents}. In: ICDAR Workshops (2019)

\bibitem{kang2024multi}
Kang, L., Tito, R., Valveny, E., Karatzas, D.: {Multi-Page Document Visual Question Answering using Self-Attention Scoring Mechanism}. arXiv preprint arXiv:2404.19024  (2024)

\bibitem{kim2022ocr}
Kim, G., Hong, T., Yim, M., Nam, J., Park, J., Yim, J., Hwang, W., Yun, S., Han, D., Park, S.: {Ocr-free document understanding transformer}. In: ECCV (2022)

\bibitem{lee2023pix2struct}
Lee, K., Joshi, M., Turc, I.R., Hu, H., Liu, F., Eisenschlos, J.M., Khandelwal, U., Shaw, P., Chang, M.W., Toutanova, K.: {Pix2struct: Screenshot parsing as pretraining for visual language understanding}. In: ICML (2023)

\bibitem{levenshtein1966binary}
Levenshtein, V.I., et~al.: {Binary codes capable of correcting deletions, insertions, and reversals}. In: Soviet physics doklady. vol.~10, pp. 707--710. Soviet Union (1966)

\bibitem{lewis2006building}
Lewis, D., Agam, G., Argamon, S., Frieder, O., Grossman, D., Heard, J.: {Building a test collection for complex document information processing}. In: SIGIR (2006)

\bibitem{lewis2019bart}
Lewis, M., Liu, Y., Goyal, N., Ghazvininejad, M., Mohamed, A., Levy, O., Stoyanov, V., Zettlemoyer, L.: {Bart: Denoising sequence-to-sequence pre-training for natural language generation, translation, and comprehension}. ACL  (2019)

\bibitem{li2022dit}
Li, J., Xu, Y., Lv, T., Cui, L., Zhang, C., Wei, F.: {Dit: Self-supervised pre-training for document image transformer}. In: ACM Multimedia (2022)

\bibitem{li2020docbank}
Li, M., Xu, Y., Cui, L., Huang, S., Wei, F., Li, Z., Zhou, M.: {DocBank: A benchmark dataset for document layout analysis}. ACL  (2020)

\bibitem{liu2024focus}
Liu, C., Wei, H., Chen, J., Kong, L., Ge, Z., Zhu, Z., Zhao, L., Sun, J., Han, C., Zhang, X.: {Focus Anywhere for Fine-grained Multi-page Document Understanding}. arXiv preprint arXiv:2405.14295  (2024)

\bibitem{liu2021swin}
Liu, Z., Lin, Y., Cao, Y., Hu, H., Wei, Y., Zhang, Z., Lin, S., Guo, B.: {Swin transformer: Hierarchical vision transformer using shifted windows}. In: ICCV (2021)

\bibitem{loshchilov2017decoupled}
Loshchilov, I., Hutter, F.: {Decoupled weight decay regularization}. arXiv preprint arXiv:1711.05101  (2017)

\bibitem{lv2023kosmos}
Lv, T., Huang, Y., Chen, J., Cui, L., Ma, S., Chang, Y., Huang, S., Wang, W., Dong, L., Luo, W., et~al.: {Kosmos-2.5: A multimodal literate model}. arXiv preprint arXiv:2309.11419  (2023)

\bibitem{mathew2021docvqa}
Mathew, M., Karatzas, D., Jawahar, C.: {Docvqa: A dataset for vqa on document images}. In: WACV (2021)

\bibitem{papineni2002bleu}
Papineni, K., Roukos, S., Ward, T., Zhu, W.J.: {Bleu: a method for automatic evaluation of machine translation}. In: ACL (2002)

\bibitem{potthast1874regesta}
Potthast, A.: {Regesta pontificum romanorum inde a 1198 ad 1304}, vol.~1. De Decker (1874)

\bibitem{raffel2020exploring}
Raffel, C., Shazeer, N., Roberts, A., Lee, K., Narang, S., Matena, M., Zhou, Y., Li, W., Liu, P.J.: {Exploring the limits of transfer learning with a unified text-to-text transformer}. Journal of Machine Learning Research  \textbf{21}(140),  1--67 (2020)

\bibitem{sun2024locr}
Sun, Y., Zhou, D., Lin, C., He, C., Ouyang, W., Zhong, H.S.: {LOCR: Location-Guided Transformer for Optical Character Recognition}. arXiv preprint arXiv:2403.02127  (2024)

\bibitem{tang2023unifying}
Tang, Z., Yang, Z., Wang, G., Fang, Y., Liu, Y., Zhu, C., Zeng, M., Zhang, C., Bansal, M.: {Unifying vision, text, and layout for universal document processing}. In: CVPR (2023)

\bibitem{taylor2022galactica}
Taylor, R., Kardas, M., Cucurull, G., Scialom, T., Hartshorn, A., Saravia, E., Poulton, A., Kerkez, V., Stojnic, R.: {Galactica: A large language model for science}. arXiv preprint arXiv:2211.09085  (2022)

\bibitem{tito2023hierarchical}
Tito, R., Karatzas, D., Valveny, E.: {Hierarchical multimodal transformers for Multipage DocVQA}. Pattern Recognit.  \textbf{144},  109834 (2023)

\bibitem{turski2023ccpdf}
Turski, M., Stanis{\l}awek, T., Kaczmarek, K., Dyda, P., Grali{\'n}ski, F.: {Ccpdf: Building a high quality corpus for visually rich documents from web crawl data}. In: ICDAR (2023)

\bibitem{van2024beyond}
Van~Landeghem, J., Biswas, S., Blaschko, M., Moens, M.F.: {Beyond document page classification: Design, datasets, and challenges}. In: WACV (2024)

\bibitem{van2023document}
Van~Landeghem, J., Tito, R., Borchmann, {\L}., Pietruszka, M., Joziak, P., Powalski, R., Jurkiewicz, D., Coustaty, M., Anckaert, B., Valveny, E., et~al.: {Document understanding dataset and evaluation (DUDE)}. In: ICCV (2023)

\bibitem{vaswani2017attention}
Vaswani, A., Shazeer, N., Parmar, N., Uszkoreit, J., Jones, L., Gomez, A.N., Kaiser, {\L}., Polosukhin, I.: {Attention is all you need}. In: NeurIPS (2017)

\bibitem{xu2020layoutlmv2}
Xu, Y., Xu, Y., Lv, T., Cui, L., Wei, F., Wang, G., Lu, Y., Florencio, D., Zhang, C., Che, W., et~al.: {Layoutlmv2: Multi-modal pre-training for visually-rich document understanding}. In: ACL (2020)

\bibitem{xu2020layoutlm}
Xu, Y., Li, M., Cui, L., Huang, S., Wei, F., Zhou, M.: {Layoutlm: Pre-training of text and layout for document image understanding}. In: SIGKDD (2020)

\bibitem{zhong2019publaynet}
Zhong, X., Tang, J., Yepes, A.J.: {Publaynet: largest dataset ever for document layout analysis}. In: ICDAR (2019)

\end{thebibliography}
\end{document}